\documentclass[11pt]{article}

\usepackage[letterpaper,top=1in,bottom=1in,left=1in,right=1in]{geometry}

\usepackage{amsmath,amssymb}
\usepackage{graphicx}
\usepackage{booktabs}
\usepackage{subcaption}
\usepackage{multirow}
\usepackage{xcolor}
\usepackage{authblk}
\usepackage{setspace}
\usepackage[hidelinks]{hyperref}
\usepackage{cleveref}
\usepackage{float} 

\usepackage{dblfloatfix}   
\usepackage{placeins}      

\newcommand{\tawss}{\mathrm{TAWSS}}
\newcommand{\mtawss}{\overline{\tawss}}

\newcommand{\fig}[2]{
  \IfFileExists{#2}{\includegraphics[width=#1]{#2}}{
    \fcolorbox{gray}{gray!12}{\parbox[c][2.6cm][c]{\dimexpr#1-2\fboxsep-2\fboxrule\relax}
    {\centering\footnotesize\ttfamily\detokenize{#2}\\[2pt]\textrm{(placeholder)}}}}}

\title{ {\bfseries Agentic AI Reasoning and Learned Sampling for Task-Aware Projection Selection in Sparse-View CT: Predicting Hemodynamics}}

\author[a,b]{Mahmoud Elhadidy}
\author[c]{Roshan M. D'Souza}
\author[a,b,d]{Amirhossein Arzani\thanks{Corresponding author: \texttt{amir.arzani@sci.utah.edu}}}

\affil[a]{Department of Mechanical Engineering, University of Utah, Salt Lake City, UT, USA}
\affil[b]{Scientific Computing and Imaging Institute, University of Utah, Salt Lake City, UT, USA}
\affil[c]{Department of Mechanical Engineering, University of Wisconsin--Milwaukee, Milwaukee, WI, USA}
\affil[d]{Department of Biomedical Engineering, University of Utah, Salt Lake City, UT, USA}

\date{}

\begin{document}

\maketitle
\begin{abstract}
\noindent
Sparse-view computed tomography (CT) reduces the number of acquired projections, but conventional sampling strategies are designed primarily for image reconstruction rather than for downstream physical inference. Here, we investigate optimal task-aware projection selection for directly predicting hemodynamics from sparse CT angiography (CTA) measurements. Using computational fluid dynamics simulations of intracranial aneurysms, we generate synthetic fan-beam CTA projections and train a Fourier neural operator (FNO) to predict two targets: the spatially resolved time-averaged wall shear stress (TAWSS) field and the spatially averaged TAWSS. The FNO was coupled with optimal input selection for simultaneous optimal sampling and training. Our goal was to investigate if agentic AI could discover optimal sampling strategies that outperform current methods. We compared fixed uniform and golden-angle sampling with three task-aware optimal selection strategies: Gumbel--Softmax (Gumbel), a Soft top-$K$ method (Soft top-$K$), and an agentic AI (Agentic) selector. The methods were compared at different view selection budgets. A detailed analysis of the AI agent was performed to understand its reasoning and performance. For spatially resolved TAWSS, Agentic and Soft top-$K$ achieve the lowest mean errors (Agentic performs slightly better) and substantially outperform fixed sampling under extreme sparsity. For spatially averaged TAWSS, optimized selection becomes important at extremely sparse acquisitions. Analysis of the selected masks reveals multiple distinct subsets concentrated around informative regions. Moreover, masks optimized for hemodynamic prediction yield poor image reconstructions, demonstrating that task-optimal acquisition differs fundamentally from reconstruction-optimal acquisition. This work suggests the possibility of designing sparse acquisitions directly around downstream physical quantities.
\end{abstract}

\bigskip

\section{Introduction}
\label{sec:intro}

Computed tomography (CT) is a well-established imaging technique. CT images are reconstructed from multiple X-ray projections~\cite{withers2021x,kato1999application,jin2026predicting}. For each slice, these projections form a sinogram, a 2D image whose axes are the projection angle and the detector position. Recovering the CT image from the sinogram is an inverse problem, classically solved with filtered backprojection (FBP), which requires densely sampled projections to give an accurate image. Modern CT scanners use iterative and model-based reconstruction methods, often initialized with FBP~\cite{hsieh2013recent,koetzier2023deep}. Acquiring many projections, however, increases the radiation dose, and high radiation dose is associated with an increase in risk of radiation-induced cancer~\cite{khatak2026radiation,smith2025projected}. Two common ways to reduce the dose are lowering the tube current of the X-ray source and acquiring fewer projections, where the latter is known as sparse-view CT~\cite{zhang2018sparse,lee2018deep}. Lowering the tube current increases
noise in the reconstructed image, while reducing the number of projections introduces imaging artifacts~\cite{frikel2013characterization}. In sparse-view CT, the views are usually acquired with a fixed, uniformly spaced pattern, since classical reconstruction theory assumes uniform angular sampling~\cite{joseph1980view,kak2001principles,zeng2022filtered}, or with
golden-angle ordering~\cite{kohler2004projection}. However, the views that are
best for image reconstruction are not necessarily optimal for a downstream prediction task~\cite{yang2025learning}, which is the focus of our study.

Prior work has addressed the reconstruction of sparse-view and low-dose CT
images~\cite{zhang2024review,chen2024deep,hsieh2013recent,koetzier2023deep,kulathilake2023review}.
Early approaches were classical, ranging from variants of FBP~\cite{enjilela2019cubic} to compressed sensing methods such as prior image constrained compressed sensing (PICCS)~\cite{chen2008prior}. Deep learning has since been applied at four levels: 1- in the sinogram domain, where the network completes or denoises the projections, 2- in the image domain, where it removes noise and artifacts from the reconstructed image, 3- in both domains jointly, known as dual-domain methods, and 4- as a direct mapping from the sparse or noisy sinogram to the final image. Across these levels, the architectures used include convolutional neural networks (CNNs)~\cite{lee2018deep}, U-Net~\cite{ernst2023sinogram,wu2019computationally},
Transformers~\cite{wang2022dudotrans,li2022ddptransformer}, neural operators~\cite{datta2026resolution}, and neural fields (implicit neural representations)~\cite{zhou2025spatiotemporal,zhou2025non}. A second direction acts at the acquisition level, learning the view mask jointly with the network during training. One group of methods treats acquisition as a sequential decision and
selects angles with reinforcement learning~\cite{shen2020learning,wang2023active,wang2025dynamic,wang2026deep}.
Alternative methods make the discrete selection differentiable and learn a fixed mask together with a reconstruction network~\cite{yang2022learning}. Most of these studies
optimize the mask and acquisition for image quality, and learning task-specific masks has received less attention~\cite{yang2025learning}.

The selection mechanisms above require either a learned policy or a differentiable relaxation of the discrete view selection. Large language models (LLMs) offer a third route. They have been used for several tasks, such as black-box optimization~\cite{yang2024large}, designing
heuristics~\cite{liu2024evolution}, mathematical discovery~\cite{romera2024mathematical}, propagating feedback as text in place
of gradients~\cite{yuksekgonul2024textgrad}, and experimental design~\cite{yuan2026unleashing,boiko2023autonomous}. Agentic artificial intelligence (AI), defined as a paradigm in which an autonomous LLM agent observes information, reasons about a goal, selects and executes actions, evaluates the outcomes, and adapts its next actions, extends the use of LLMs further. Agentic AI has recently been used across scientific fields, including the discovery of numerical evolution algorithms~\cite{toscano2026graft}, fluid flow modeling~\cite{lee2026agentic}, and scientific machine learning~\cite{jiang2026agenticsciml}. In computational imaging, agents have been used to support CT reconstruction~\cite{maier2026agentic}, where an LLM-based agent was shown to implement, tune, and benchmark multiple CT reconstruction methods under a fixed compute budget and a prescribed acquisition geometry. Their agent therefore operates at the reconstruction level, where the projection measurements are fixed and the question of which projections should be acquired is not addressed. Agentic methods have also been used to select magnetic resonance sequences~\cite{zaiss2026agentic}.

While previous work has addressed task-specific masks, the chosen tasks were
classification~\cite{yang2025learning}, neglecting more complex tasks such as hemodynamic prediction, which is important for studying cardiovascular disease~\cite{meng2013high,zhou2017association,mahmoudi2021story}. This comes despite increased efforts to infer hemodynamics such as velocity and wall shear stress (WSS) from CT angiography (CTA) or concentration data, either using classical methods~\cite{lardo2015estimating,eslami2015computational} or deep learning and physics-informed neural networks~\cite{raissi2020hidden,elhadidy2026sle,elhadidy2026wall}.
This raises two questions. Can the selected sinogram views be optimized for hemodynamic prediction rather than for image quality? And can agentic AI reasoning find the best subset of views, and how does this compare with other learning-based optimal selection methods?

These questions motivate us to study the feasibility of optimizing the sinogram view selection for hemodynamic prediction, and of using agentic AI reasoning to achieve this task. To the best of our knowledge, this is the first work to address task-aware sparse-view CTA selection for hemodynamic prediction, and the first to use agentic AI for view selection in sparse-view CT. Our contributions are as follows:

\begin{itemize}
    \item We develop a task-aware sparse-view CTA framework in which the view
    mask is selected under a fixed view budget for hemodynamic prediction.

    \item We introduce an agentic selector, in which an LLM-based agent
    proposes, evaluates, and refines candidate view sets, and compare its performance
    against learned and fixed selection methods under the same budgets.

    \item We provide a detailed analysis of the agentic AI selector's reasoning,
    hypotheses, and mask evaluations.

    \item We consider two downstream objectives: predicting the spatially
    averaged time-averaged WSS (TAWSS), and predicting the spatially resolved TAWSS field.
\end{itemize}

Our contributions allow the acquisition to be designed around a downstream physical quantity rather than image quality. The framework is not specific to hemodynamics, and the same selection mechanisms can be applied to other tasks in which the target is derived from the measured projections. Because the agentic selector searches directly over discrete
view subsets, it can also accommodate acquisition constraints that are difficult to express in a differentiable form and our study represents an early application of LLM reasoning for optimal sensor placement, which is a problem of broader interest in scientific modeling~\cite{brunton2013optimal,semaan2017optimal,nav2026hierarchical}.

\section{Methods}
\label{sec:method}

\begin{figure}[htbp]
  \centering
  \includegraphics[width=\textwidth]{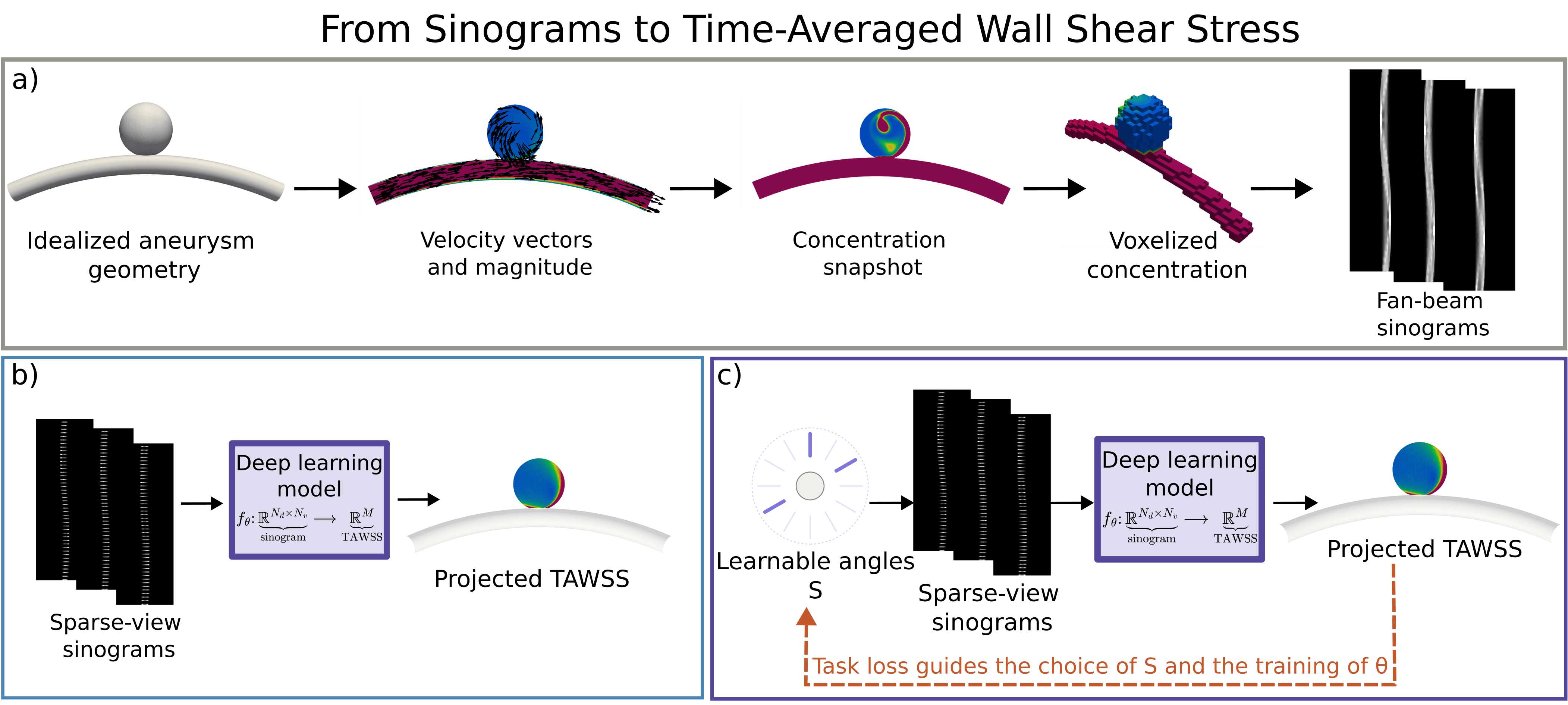}
  \caption{Overview of the study. \textbf{(a)} Data generation: an idealized
  aneurysm geometry gives a pulsatile velocity field, which drives an
  advection--diffusion simulation of contrast concentration. The
  concentration field is voxelized and fan-beam sinograms are synthetically generated. The training dataset contained distinct geometry and flow rates.
  \textbf{(b)} Prediction from a fixed sparse view set: the predictor
  $f_\theta:\mathbb{R}^{N_d\times N_v}\rightarrow\mathbb{R}^{M}$ maps the
  sparse-view sinogram directly to the projected time-averaged wall shear
  stress (TAWSS) on the vessel wall, where $M$ is the number of predicted
  outputs as introduced in Sec.~\ref{ssec:FNO}. \textbf{(c)} Prediction with learned view selection: the acquired
  subset $\mathcal{S}$ of the $N_v$ candidate views is optimized together with
  the predictor model $f_\theta$, guided by the task loss.}
  \label{fig:overview}
\end{figure}

\subsection{Problem Formulation}
\label{ssec:formulation}

Let $\mathbf{C}^{(t)}\in\mathbb{R}^{N_x\times N_y\times N_z}$ denote the
voxelized contrast concentration volume of a vessel segment at time frame
$t\in\{1,\ldots,N_t\}$, i.e., a snapshot of the concentration at one of $N_t$
time instants within one cardiac cycle, where each entry is the contrast
concentration in one voxel, and $N_x$, $N_y$ and $N_z$ are the numbers of voxels along the
three grid directions. In general these three dimensions can differ. In this
work, we use a cubic grid with $N_x=N_y=N_z$. The concentration is not periodic and follows the injected bolus and its washout. The
volume is scanned slice by slice along the $y$ direction, so each slice lies
in the $x$--$z$ plane and $N_y$ is the number of slices. For time frame $t$ and
slice $y\in\{1,\ldots,N_y\}$, with in-plane concentration
$\mathbf{C}^{(t)}_y\in\mathbb{R}^{N_x\times N_z}$, let $\mathcal{R}_\phi$
denote the 2D fan-beam forward projection operator at gantry angle $\phi$.
A full acquisition samples the candidate angle set
$\Phi=\{\phi_1,\ldots,\phi_{N_v}\}$, giving the slice sinogram
\begin{equation}
\mathbf{X}^{(t)}_y
=
\mathcal{R}_\Phi\mathbf{C}^{(t)}_y
\in\mathbb{R}^{N_d\times N_v},
\label{eq:slice_sinogram}
\end{equation}
where $N_v$ is the number of candidate views and $N_d$ is the number of
detector elements.

Let $\boldsymbol{\tau}_w(\mathbf{p},t)$ denote the instantaneous WSS vector (hemodynamic target in this work) at
point $\mathbf{p}$ on the vessel surface $\Gamma$ and time $t$ over one
cardiac cycle of duration $T$. We consider two targets:
\begin{align}
  \label{eq:t1}
  \text{(T1)} \quad & \tawss(\mathbf{p}) = \tfrac{1}{T}\!\int_0^T \!\!
    \|\boldsymbol{\tau}_w(\mathbf{p},t)\|\,dt , \\
  \label{eq:t2}
  \text{(T2)} \quad & \mtawss = \tfrac{1}{|\Gamma|}\!\int_\Gamma \!
    \tawss(\mathbf{p})\,d\Gamma ,
\end{align}
where $\|\cdot\|$ is the Euclidean norm and $|\Gamma|$ is the surface area of
the vessel wall (here, the aneurysm sac). T1 is the spatially resolved TAWSS, a scalar field over
$\Gamma$, whereas T2 is the spatially averaged TAWSS, a single scalar per
case. Our aim is to map the sinogram measurements directly to these targets,
without reconstructing an image. An overview of the problem formulation is shown in Fig.~\ref{fig:overview}.

The network does not receive the slice sinograms individually. Instead, they
are averaged over all time frames of one cardiac cycle and over a set
$\mathcal{Y}\subseteq\{1,\ldots,N_y\}$ of central slices along the $y$
direction that contain the contrast, giving a single network input
\begin{equation}
\mathbf{X}
=
\frac{1}{N_t\,|\mathcal{Y}|}
\sum_{t=1}^{N_t}\sum_{y\in\mathcal{Y}}\mathbf{X}^{(t)}_y
=
[\mathbf{x}_1,\ldots,\mathbf{x}_{N_v}]
\in\mathbb{R}^{N_d\times N_v},
\label{eq:sinogram}
\end{equation}
where $\mathbf{x}_j\in\mathbb{R}^{N_d}$ is the column of aggregated
measurements at view $j$, and $N_t$ and $\mathcal{Y}$ are specified in
Sec.~\ref{ssec:data}. The aggregation matches the structure of the targets: TAWSS is itself
averaged over the cardiac cycle and defined over the whole sac surface, so
each case has a single target, either one field or one scalar, rather than
one per time frame or per slice. Averaging the inputs over time frames and slices
therefore gives one measurement per case to pair with that single target. Because the forward projection is linear,
Eq.~\eqref{eq:sinogram} equals the fan-beam sinogram of the frame- and
slice-averaged concentration distribution, so the predictor receives one
compressed 2D measurement per case rather than the full stack. Combining measurements over time and across neighboring slices is common in dynamic CT~\cite{sliwicka2023dynamic} and in multislice network inputs~\cite{vu2020evaluation}. This aggregation is
used only to form the network input.  The per-slice, per-frame sinograms remain available for the image reconstructions pursued in Sec.~\ref{ssec:CT_reconstruct}. Since every slice and
time frame is acquired at the same gantry angles, a single view selection applies to all of them. Sparse-view acquisition keeps a subset $\mathcal{S}\subseteq\{1,\ldots,N_v\}$ of $|\mathcal{S}|=K\ll N_v$ views, encoded by the binary mask $\mathbf{b}\in\{0,1\}^{N_v}$ with
\begin{equation}
b_j=\mathbb{I}[j\in\mathcal{S}],
\qquad
\|\mathbf{b}\|_0=K,
\qquad
\widetilde{\mathbf{X}}=\mathbf{X}\odot\mathbf{b},
\label{eq:mask}
\end{equation}
where $\mathbb{I}[\cdot]$ is the indicator function, $K$ is the view budget, $\odot$ multiplies each column $\mathbf{x}_j$ by the corresponding mask entry $b_j$, and $\widetilde{\mathbf{X}}$ is the masked sinogram. In general, $\widetilde{\mathbf{X}}$ may first be passed through a reconstruction or preprocessing operator. In this work, we directly map the masked sinogram to the hemodynamic target using a neural-network predictor $f_{\theta}$, where $f_{\theta}$ denotes the parameterized mapping from the masked sinogram $\widetilde{\mathbf{X}}$ to the predicted target
$\widehat{y}^{(q)}$, and $\theta$ denotes the trainable parameters of the predictor:
\begin{equation}
\widehat{y}^{(q)}
=
f_{\theta}\!\left(\widetilde{\mathbf{X}}\right).
\label{eq:predictor}
\end{equation}
The goal is therefore to jointly determine an informative view mask
$\mathbf{b}$ and learn the predictor parameters $\theta$ for the downstream hemodynamic prediction task.

Given a task $q\in\{\mathrm{T1},\mathrm{T2}\}$ with target $y^{(q)}$ and loss $\mathcal{L}^{(q)}$, we select the view mask and train the predictor to solve
\begin{equation}
  \min_{\mathbf{b},\,\theta} \;
  \mathbb{E}_{\mathbf{C}\sim\mathcal{D}}
  \Big[ \mathcal{L}^{(q)}\!\big( f_\theta(\mathbf{X}\odot\mathbf{b}),\,
  y^{(q)} \big) \Big]
  \quad \text{s.t.} \quad \|\mathbf{b}\|_0 = K,
  \label{eq:objective}
\end{equation}
where $\mathcal{D}$ is the distribution of concentration volumes. The mask is therefore optimized for hemodynamic prediction accuracy rather than image reconstruction fidelity. Eq.~\eqref{eq:objective} is solved separately for each task ($\mathrm{T1}$, $\mathrm{T2}$), allowing the selected mask $\mathbf{b}^{(q)}$ to be optimized for the desired task. All methods compared use exactly $K$ distinct views as the budget for a consistent comparison.

\subsection{Dataset and CTA Simulation}
\label{ssec:data}

Our dataset consisted of 220 computational fluid dynamics (CFD) simulations of an idealized sidewall aneurysm, based on the configuration used in our previous study~\cite{elhadidy2026sle}, where the geometry (aneurysm neck size, aneurysm width, and parent artery curvature) and the inflow boundary condition were varied across the CFD simulations. To evaluate the proposed approach under more challenging flow conditions, we extended the Reynolds number range to
260--570, beyond the relatively lower values considered previously. Contrast
transport was modeled by solving the advection--diffusion equation using a passive scalar as the contrast surrogate, with a trapezoidal inlet concentration profile and washout during the final cardiac cycle. Five time frames ($N_t=5$) were sampled at quarter-cycle spacing, mimicking a gated multi-phase acquisition, and all five are aggregated into the network input (Eq.~\eqref{eq:sinogram}).

Each frame was resampled from the unstructured CFD mesh onto a common Cartesian grid of $64^3$ voxels ($N_x=N_y=N_z=64$), giving the volume $\mathbf{C}^{(t)}$ of Sec.~\ref{ssec:formulation}. Each voxel holds the mean concentration of the CFD nodes inside it. The acquisition was then modeled as a stack of $N_y=64$ independent 2D axial scans, one per slice along the $y$ direction, running from the parent artery through the aneurysm sac. We used a fan-beam geometry to approximate the in-plane acquisition geometry of clinical multidetector CT scanners used for CTA of intracranial aneurysms~\cite{guo2026sinogram}. Each slice $\mathbf{C}^{(t)}_y$ is a $64\times64$ grid and is projected separately. In this 2D fan-beam model, every ray lies
entirely within the plane of its own slice, so the sinogram of a slice depends only on the concentration in that slice and receives no information from neighboring slices. This differs from cone-beam CT, in which the rays diverge across several slices and each detector reading mixes information from a 3D region.

Projections were computed with the ASTRA Toolbox~\cite{palenstijn2013astra}, which provides the discretized tomographic forward operators. We used its \texttt{fanflat} geometry (2D fan beam, flat detector) with source-to-isocenter
and source-to-detector distances of 600 and 1100~mm, and a per-case detector pitch. For each slice, the forward projection returns, for every detector element, the integral of the concentration along the straight ray from the X-ray source to that element~\cite{kak2001principles}:
\begin{equation}
p^{(t)}_y(\phi,d)
=
\int_{0}^{D_{\mathrm{sd}}}
\mathbf{C}^{(t)}_y\!\left(\mathbf{a}_\phi+s\,\mathbf{u}_{\phi,d}\right)
\mathrm{d}s,
\label{eq:line_integral}
\end{equation}
where $\mathbf{a}_\phi$ is the source position at gantry angle $\phi$, $\mathbf{u}_{\phi,d}$ is the unit vector pointing from the source toward detector element $d$, $s$ is the distance traveled along the ray, and $D_{\mathrm{sd}}=1100$~mm is the source-to-detector distance. Collecting $p^{(t)}_y(\phi,d)$ over all angles $\phi\in\Phi$ and detector elements $d=1,\ldots,N_d$ gives the slice sinogram $\mathbf{X}^{(t)}_y$ of Eq.~\eqref{eq:slice_sinogram}. We acquired $N_v=180$ views at $2^\circ$ increments, with $N_d=64$ detector elements per view. The forward model is idealized: monochromatic, noise-free, and without beam hardening, scatter, or detector blur. Measurement noise is added separately in Sec.~\ref{ssec:noise_robustness}. We further assume that the contrast is quasi-static while the projections of one time frame are acquired, although the gantry rotates during a real scan. This assumption follows previous work~\cite{lardo2015estimating,eslami2015computational,thombre2026vasto} and accounting for the change in concentration during each full acquisition is left for future work.

The per-slice sinograms were aggregated into the single $N_d\times N_v$
sinogram of Eq.~\eqref{eq:sinogram} by averaging over all $N_t=5$ frames of
one cardiac cycle and over the 32 central slices along the $y$ direction,
so that $|\mathcal{Y}|=32$ of the $N_y=64$ slices enter the input. The
remaining slices lie outside the vessel and are almost entirely zero, so they
carry no contrast information. Restricting the input to the region that
contains the object is a standard preprocessing step in medical image
analysis~\cite{isensee2021nnu,christ2016automatic,roth2018application,zhou2017fixed},
and using a selected set of neighboring slices, rather than the full volume,
as network input is common
practice~\cite{vu2020evaluation,han2017automatic,li2018h}. No
single frame is used on its own at any stage of training or evaluating the
predictor. Since all slices share the same gantry angles, one view selection
applies to the whole stack. The corresponding TAWSS fields were computed from the CFD solution on the curved surface mesh of the aneurysm sac and were then projected onto a $64\times64$ planar grid using the procedure described in our previous study~\cite{elhadidy2026sle}. This projection only serves to represent the wall field as a 2D image for the FNO output (each grid value corresponds to a point on the curved wall).
Models were trained to predict the two targets T1 and T2 from the
aggregated sinogram.

\subsection{FNO Architecture for Wall Shear Stress Prediction}
\label{ssec:FNO}

Predicting T1 or T2 from sinograms is formulated as a supervised operator learning
problem. We adopt a Fourier neural operator (FNO)
architecture~\cite{li2020fourier}, motivated by its effectiveness in learning
mappings between function spaces and approximating solution operators for
partial differential equations~\cite{li2020fourier,kovachki2023neural,li2023fourier}, as well as
its application to hemodynamic inference~\cite{flouris2025localized,elhadidy2026sle}.

We define the neural network $f_\theta$ as a two-dimensional FNO. The network
takes the single-channel masked sinogram $\widetilde{\mathbf{X}}\in\mathbb{R}^{N_d\times N_v}$ as input and uses four Fourier layers with 32 hidden channels. Retaining 16 Fourier modes along each dimension yields the retained spectral coefficients. For T1, the network predicts the spatially resolved TAWSS on a $64\times64$ grid. For T2, the extracted features are pooled and mapped to a single scalar.

A separate predictor is trained for each target using the mean squared error
(MSE) loss,
\begin{equation}
\mathcal{L}_{\mathrm{MSE}}
=
\frac{1}{M}
\sum_{m=1}^{M}
\left(
\widehat{y}_m-y_m
\right)^2,
\label{eq:loss}
\end{equation}
where $\widehat{y}_m$ and $y_m$ are the predicted and reference values of the $m$-th output, and $M$ is the number of outputs, with $M=64\times64$ for T1 and $M=1$ for T2. 

\subsection{View Selection}
\label{ssec:views}

\begin{figure}[htbp]
  \centering
  \includegraphics[width=\textwidth]{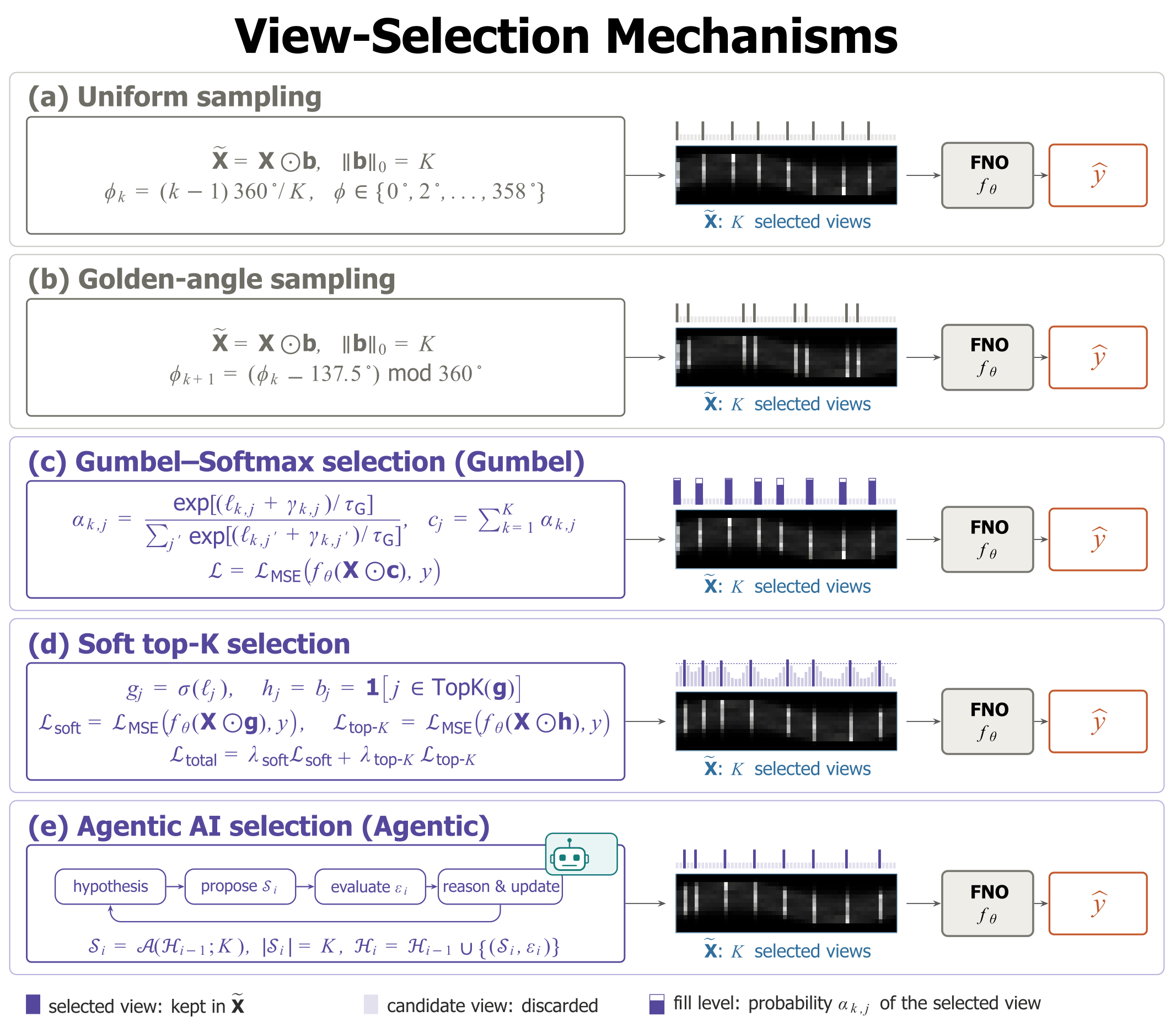}
  \caption{Five mechanisms for choosing which $K$ views of the $N_v$ candidates are selected. Each row gives the defining equations on the left, the
  resulting selection in the center, and a similar predictor $f_\theta$,
  implemented as an FNO, on the right. 
  \textbf{(a)} Uniform sampling: $K$ equispaced angles, fixed before training.
  \textbf{(b)} Golden-angle sampling: fixed $137.5^{\circ}$ increments.
  \textbf{(c)} Gumbel--Softmax selection (Gumbel).
  \textbf{(d)} Soft top-$K$ selection.
  \textbf{(e)} Agentic AI selection (Agentic).}
  \label{fig:mechanisms}
\end{figure}


Three view selection approaches were evaluated: Gumbel--Softmax selection (\textit{Gumbel}), Soft top-$K$ selection (\textit{Soft top-$K$}), 
and agentic AI selection (\textit{Agentic}). Two fixed-sampling baselines
were also considered: uniform angular sampling and golden-angle
sampling~\cite{yang2022learning}. These fixed view baselines select $K$ views without optimizing their angular locations for the downstream prediction task. Fig.~\ref{fig:mechanisms} illustrates the three mechanisms together with the two baselines. Every method produces a subset $\mathcal{S}$ and mask $\mathbf{b}$ as in Eq.~\eqref{eq:mask}, and the methods differ only in how $\mathcal{S}$ is chosen. For Gumbel and Soft top-$K$, the selection parameters and the predictor are initially optimized together. At a prescribed training epoch, the selection is frozen and the unselected views and their associated weights are pruned. Training then continues for the predictor alone, with the selected views held fixed for further fine-tuning of the prediction.

\subsubsection{Gumbel--Softmax Selection (Gumbel)}
\label{sssec:gumbel}

The first learned mechanism is based on the Gumbel--Softmax relaxation,
which has previously been used for differentiable feature and view
selection~\cite{maddison2017a,balin2019concrete,jang2017categorical}. Each of
the $K$ selection units, indexed by $k=1,\ldots,K$, contains a trainable
logit $\ell_{k,j}$ for every candidate view $j$. During joint training,
Gumbel noise is sampled according to
\begin{equation}
\gamma_{k,j}
=
-\log\left[-\log\left(u_{k,j}\right)\right],
\qquad
u_{k,j}\sim\mathrm{Uniform}(0,1),
\label{eq:gumbel_noise}
\end{equation}
where $\gamma_{k,j}$ is the noise added to the logit of view $j$ in unit $k$
and $u_{k,j}$ is a uniform random variate. The corresponding soft selection
weights are
\begin{equation}
\alpha_{k,j}
=
\frac{
\exp\left[
\left(\ell_{k,j}+\gamma_{k,j}\right)/\tau_{\mathrm{G}}
\right]
}{
\displaystyle\sum_{j'=1}^{N_v}
\exp\left[
\left(\ell_{k,j'}+\gamma_{k,j'}\right)/\tau_{\mathrm{G}}
\right]
},
\label{eq:gumbel_softmax}
\end{equation}
where $\alpha_{k,j}\in(0,1)$ is the relaxed weight assigned to view $j$ by
unit $k$, with $\sum_{j}\alpha_{k,j}=1$, and $\tau_{\mathrm{G}}$ is a
temperature hyperparameter. A high temperature produces a smooth distribution over the candidate views, whereas decreasing $\tau_{\mathrm{G}}$ makes the selection vector approach a one-hot representation. The $K$ units are summed into a single gate per view, which scales the corresponding column of the full sinogram:
\begin{equation}
c_j
=
\sum_{k=1}^{K}\alpha_{k,j},
\qquad
\widetilde{\mathbf{X}}
=
\mathbf{X}\odot\mathbf{c}
\in\mathbb{R}^{N_d\times N_v},
\label{eq:selected_input}
\end{equation}
where $\mathbf{c}=[c_1,\ldots,c_{N_v}]$ is the per-view gate. The predictor
therefore receives a sinogram of the same size as for the other methods, with
each view kept at its own angular position and weighted by its gate. This
continuous relaxation allows the selection parameters and the predictor to be
optimized jointly using backpropagation.

At the prescribed pruning epoch, the selector is frozen and a subset
$\mathcal{S}$ of exactly $K$ distinct views is constructed. First, the view
with the largest learned logit is identified for each unit:
\begin{equation}
j_k^*
=
\mathop{\mathrm{arg\,max}}_{1\leq j\leq N_v}
\ell_{k,j},
\label{eq:hard_view_selection}
\end{equation}
where $j_k^*$ is the preferred view of unit $k$. Duplicate selections are
then removed. If fewer than $K$ distinct views remain, the remaining budget is
filled using previously unselected views. For the spatially averaged TAWSS,
these are ranked by their importance relative to the preferred view of each
unit:
\begin{equation}
r_{k,j}
=
\frac{\pi_{k,j}}{\pi_{k,j_k^*}},
\label{eq:relative_view_importance}
\end{equation}
where $r_{k,j}$ is the relative importance of view $j$ in unit $k$ and
$\pi_{k,j}$ is the selection probability of view $j$ in unit $k$, given by
Eq.~\eqref{eq:gumbel_softmax} with $\gamma_{k,j}=0$. For the spatially
resolved TAWSS, the views are ranked directly by the absolute probability
$\pi_{k,j}$. Each ranking was chosen because it gave the best results for its
target. In both cases, views are added in decreasing order of the ranking
score until $|\mathcal{S}|=K$, and the mask $\mathbf{b}$ follows from
Eq.~\eqref{eq:mask}. Unselected views are then pruned, and fine-tuning
continues for the predictor alone on
$\widetilde{\mathbf{X}}=\mathbf{X}\odot\mathbf{b}$, without further Gumbel
sampling.

\subsubsection{Soft Top-$K$ Selection}
\label{sssec:soft_topk}

The second learned view selection mechanism was adapted from the fractal
autoencoder (FAE) of Wu and Cheng~\cite{wu2021fractal}. While the original
formulation uses learnable feature weights $w_j$, we assign each candidate
view a learnable sigmoid gate. This parameterization bounds the weights
within $(0,1)$, allowing smooth optimization while controlling view
suppression or retention without amplifying the input measurements. Similar
sigmoid-based feature weighting is used in squeeze-and-excitation
networks~\cite{hu2018squeeze}. The gates are defined as
\begin{equation}
g_j
=
\sigma(\ell_j)
=
\frac{1}{1+\exp(-\ell_j)},
\qquad
j=1,\ldots,N_v,
\label{eq:sigmoid_gate}
\end{equation}
where $g_j\in(0,1)$ is the gate of view $j$, $\ell_j$ is the corresponding
trainable logit, and $\sigma(\cdot)$ is the logistic sigmoid.

We refer to this approach as Soft top-$K$ to reflect its two complementary
selection mechanisms: soft weighting of all views through continuous gates,
and hard selection of the $K$ highest-weighted views, following the
feature-selection mechanism of FAE~\cite{wu2021fractal}. The name describes
our adaptation of these mechanisms to projection view selection with an FNO
predictor, rather than a separately established algorithm.

Two complementary branches are used during training. Both branches see the
same sinogram and the same gates, and they differ only in how many views are
kept. The soft branch weights every view by its gate, whereas the top-$K$
branch keeps only the $K$ views with the largest gates and discards the rest,
\begin{equation}
h_j
=
\begin{cases}
1, & j\in\operatorname{TopK}(\mathbf{g}),\\
0, & \text{otherwise},
\end{cases}
\qquad
\mathcal{S}=\operatorname{TopK}(\mathbf{g}),
\label{eq:weighted_topk_gate}
\end{equation}
where $h_j=b_j$ is the binary mask of Eq.~\eqref{eq:mask} for this subset,
$\mathbf{g}=[g_1,\ldots,g_{N_v}]$ collects the gates, and
$\operatorname{TopK}(\mathbf{g})$ returns the indices of the $K$ largest gate
values. Since $h_j$ is piecewise constant, the top-$K$ branch is made
differentiable with a straight-through estimator~\cite{bengio2013estimating}: the forward pass uses the
binary values above, while the backward pass uses the gradient of $g_j$, so
the logits receive a gradient from both branches. The inputs to the two
branches are
\begin{equation}
\widetilde{\mathbf{X}}_{\mathrm{soft}}
=
\mathbf{X}\odot\mathbf{g},
\qquad
\widetilde{\mathbf{X}}_{\mathrm{top}\text{-}K}
=
\mathbf{X}\odot\mathbf{h},
\label{eq:soft_topk_inputs}
\end{equation}
where $\mathbf{h}=[h_1,\ldots,h_{N_v}]$ collects the top-$K$ weights, and
$\widetilde{\mathbf{X}}_{\mathrm{soft}}$ and
$\widetilde{\mathbf{X}}_{\mathrm{top}\text{-}K}$ are the inputs passed to the
predictor by the two branches.

At each training epoch, $\mathcal{S}$ is recomputed from the current gate
values, so the selected subset can change during training until the selector
is frozen. The network is optimized using the combined objective
\begin{equation}
\mathcal{L}_{\mathrm{total}}
=
\lambda_{\mathrm{soft}}
\mathcal{L}_{\mathrm{soft}}
+
\lambda_{\mathrm{top}\text{-}K}
\mathcal{L}_{\mathrm{top}\text{-}K},
\label{eq:soft_topk_loss}
\end{equation}
where $\mathcal{L}_{\mathrm{soft}}$ and $\mathcal{L}_{\mathrm{top}\text{-}K}$
are the losses evaluated on
$\widetilde{\mathbf{X}}_{\mathrm{soft}}$ and
$\widetilde{\mathbf{X}}_{\mathrm{top}\text{-}K}$, respectively, and
$\lambda_{\mathrm{soft}}$ and $\lambda_{\mathrm{top}\text{-}K}$ weight their
contributions.

Training proceeds in two phases. In the first phase, both branches are
evaluated at every step and a single predictor is trained on their combined
loss, with $\lambda_{\mathrm{soft}}=1$ and
$\lambda_{\mathrm{top}\text{-}K}=50$. The soft branch keeps a gradient on
every view, including those currently outside the top-$K$ set, so the
selection can still change. The top-$K$ branch is weighted more heavily
because it matches the acquisition that will be deployed. At the freezing point, the current top-$K$ set becomes the final subset $\mathcal{S}$, the kept gates are set to one, the logits stop receiving updates, and the unselected views are pruned. In the second phase, the soft branch is discarded and only the predictor is trained, on the binary masked sinogram $\widetilde{\mathbf{X}}=\mathbf{X}\odot\mathbf{b}$.
Prediction at test time therefore comes from the top-$K$ branch alone, and
the soft branch serves only to keep the selection trainable during the first
phase.

Although the original formulation included an $L_1$ penalty on the feature
weights, we omit this term because the view budget is already enforced by the
top-$K$ branch. Our experiments also showed that adding $L_1$ regularization
provided no additional benefit.

\subsubsection{Agentic AI Selection}
\label{sssec:agentic_selection}

The third view selection mechanism employs agentic AI (Agentic) as the optimal selector. Rather than optimizing continuous selection parameters, the agent searches directly over discrete view subsets. It iteratively proposes a candidate subset, trains the predictor using that subset, evaluates its performance on the validation set, and uses the resulting validation error and LLM reasoning to guide its subsequent proposals. Each candidate subset satisfies the view budget:
\begin{equation}
\mathcal{S}_i \subseteq \{1,\ldots,N_v\},
\qquad
|\mathcal{S}_i|=K,
\label{eq:agentic_subset}
\end{equation}
where $\mathcal{S}_i$ is the subset proposed at search trial $i$. After each
trial, the resulting validation error, together with the history of
previously evaluated subsets, is provided to the agent:
\begin{equation}
\mathcal{S}_i
=
\mathcal{A}(\mathcal{H}_{i-1};K),
\qquad
\mathcal{H}_{i-1}
=
\{(\mathcal{S}_{1},\varepsilon_{1}),\ldots,
(\mathcal{S}_{i-1},\varepsilon_{i-1})\},
\label{eq:agentic_search}
\end{equation}
where $\mathcal{A}$ denotes the agentic decision-making process,
$\mathcal{H}_{i-1}$ is the search history, and $\varepsilon_{i'}$ is the
validation Rel-$L_2$ error of trial $i'$. The agent uses this history to
reason about subsequent modifications to the selected views.

The number of search trials is not fixed in advance. Instead, the agent
continues proposing and evaluating new subsets until it determines that
further exploration is unlikely to improve the solution. It then selects the
most promising subset $\mathcal{S}^{*}$ from the candidates and validation
results observed during its search. The final predictor is trained using this
fixed subset for the full training schedule and is then evaluated on the test
set. The corresponding input is
\begin{equation}
\widetilde{\mathbf{X}}
=
\mathbf{X}\odot\mathbf{b}^{*},
\qquad
b_j^{*}
=
\mathbb{I}[j\in\mathcal{S}^{*}],
\label{eq:agentic_final_mask}
\end{equation}
where $\mathcal{S}^{*}$ is the final agent-selected subset and $\mathbf{b}^{*}$ the
corresponding mask of Eq.~\eqref{eq:mask}. 

Therefore, in this method, there is no joint gradient-based optimization of the selector and predictor. The predictor parameters are optimized during each search trial, while the agent modifies the discrete view subset based on the validation feedback from its previous trials. After the final subset is selected, only the predictor is trained during the full training run. Claude Code provides the agentic framework for coordinating the search, training, evaluation, and reasoning process. Opus 5.0 model was used as our agentic LLM model.

To quantify the variability in the agentic search, five independent agents
are used for every combination of task and view budget. Each agent starts
with a fresh context and has no access to the search history, selected subsets, reasoning, or results of the other agents. It proposes and evaluates candidate subsets until it decides that further exploration is unlikely to help, so the number of trials differs from one agent to another. With two targets, five view budgets $K\in\{90,45,15,2,1\}$, and five agents per combination, this gives 50 independent searches. On average, each agent performed around 8 trials and a total of 390 trials was conducted.

For every trial, the agent logs the proposed subset, validation error, and a concise rationale for its decision. This provides an auditable record of the search trajectory and allows the reasoning and view selection behavior to be analyzed across independent agents. Analysis of the behavior is presented in the results.

\subsection{Training Protocol and Hyperparameters}
\label{ssec:training}

All methods used the same FNO predictor with four Fourier layers, 32 hidden channels, and $16\times16$ retained Fourier modes. The models were trained using AdamW for 3000 epochs with a learning rate of $3\times10^{-3}$, weight decay of $10^{-4}$, batch size 16, cosine learning-rate annealing, and MSE loss. Inputs were normalized by the 99th percentile of training sinogram values and targets by the maximum training target value. The dataset contained 200 training and 20 test cases (in the case of Agentic, the validation set is a subset of the training set), and results were reported over five random seeds and, for Agentic it was reported over five independent runs. For Soft top-$K$ and Gumbel, the view selector was optimized jointly with the predictor for the first 500 epochs. The resulting exact-$K$ mask was then frozen, and the predictor alone was trained for the remaining 2500 epochs. Soft top-$K$ used $\lambda_{\mathrm{soft}}=1$ and $\lambda_{\mathrm{top}\text{-}K}=50$. Gumbel used a temperature annealed from $2.0$ to $0.01$ over the first 500 epochs. Uniform and golden-angle sampling used fixed masks throughout all 3000 epochs.

\section{Results}
\label{sec:results}

The described methods were compared at view budgets of \(K\in\{90,45,15,2,1\}\). For \(K=1\), both the uniform and golden-angle baselines select the projection at \(\phi=0^\circ\), and therefore use the same single-view mask. Their small differences arise from non-deterministic GPU arithmetic during training (runs are seeded but not bitwise reproducible). Prediction accuracy is quantified using the mean relative \(L_2\) error (Rel-\(L_2\)) over the test set:
\begin{equation}
\mathrm{Rel}\text{-}L_2
=
\frac{1}{N_{\mathrm{test}}}
\sum_{i=1}^{N_{\mathrm{test}}}
\frac{\left\|\widehat{\mathbf{y}}_i-\mathbf{y}_i\right\|_2}
     {\left\|\mathbf{y}_i\right\|_2},
\label{eq:test_rel_l2}
\end{equation}
where \(N_{\mathrm{test}}\) is the number of test samples, and
\(\widehat{\mathbf{y}}_i\) and \(\mathbf{y}_i\) denote the predicted and
reference outputs, respectively. For spatially averaged TAWSS, the
outputs are scalars, and the norm reduces to the absolute value.

Results are reported as mean \(\pm\) standard deviation across five
runs for each combination of selection method, view budget, and
prediction target. The random seeds control neural network weight
initialization, training data shuffling, and the train/test split.
The same five seeds are used across methods to enable consistent
comparisons. For the agentic AI approach, the five runs correspond to five independent agentic searches, each starting with a fresh agent context.

\subsection{Spatially Resolved TAWSS}
\label{ssec:field_results}

\begin{figure}[htbp]
  \centering
  \includegraphics[width=\textwidth]{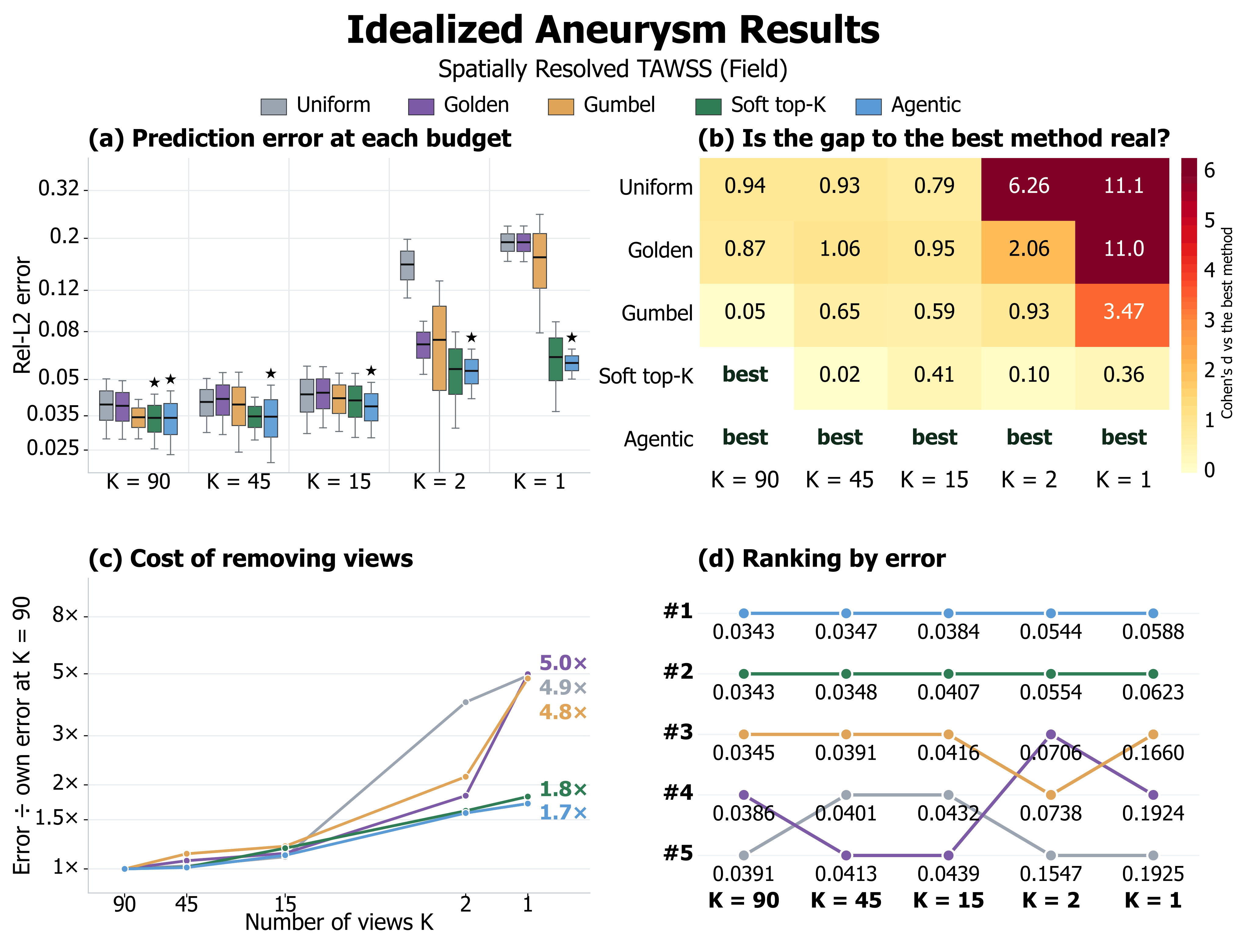}
  \caption{Prediction of spatially resolved TAWSS at different view budgets $K$. 
    \textbf{(a)} Test Rel-$L_2$ error across five runs.
    Boxes span the mean $\pm 1$ standard deviation,
    center lines indicate the mean, and whiskers span the mean
    $\pm 2$ standard deviations. Stars mark the lowest
    mean error at each budget.
    \textbf{(b)} Gap to the best method at each budget, expressed as
    Cohen's $d$: the difference in mean error divided by the pooled
    standard deviation of the two methods,
    $d = (\bar{\varepsilon}_i - \bar{\varepsilon}_{\mathrm{best}}) /
    \sqrt{(s_i^2 + s_{\mathrm{best}}^2)/2}$. The best method at each
    budget is marked \emph{best}.
    \textbf{(c)} Mean error normalized by each method's mean error
    at $K=90$, showing the relative increase as the view budget
    decreases.
    \textbf{(d)} Methods ranked by mean error at each budget, with rank~1 indicating the lowest error. At $K=90$, Agentic and Soft top-$K$ are tied. Agentic is shown as rank~1 solely for visualization purposes.}
  \label{fig:field}
\end{figure}

We evaluate how well each method predicts spatially resolved TAWSS as
the number of views is reduced (Fig.~\ref{fig:field},
Table~\ref{tab:idealized-field-results}). Agentic and Soft top-$K$ hold the lowest
mean error at every budget, but the size of that advantage depends strongly on
how many views remain. At $K = 90$, all five methods lie within one standard
deviation of each other (Fig.~\ref{fig:field}a, Table~\ref{tab:idealized-field-results}), so a dense budget does not meaningfully reward task-aware selection. The
same holds down to $K = 15$, where the errors range from $0.0384$ (Agentic) to
$0.0439$ (Golden) and no method's error increases by more than about $21\%$ relative to its $K = 90$ value.

The distinctions between the selection methods become more pronounced below $K = 15$ (Fig.~\ref{fig:field}b). At $K = 2$, Agentic and Soft top-$K$ reach $0.0544$ and $0.0554$ error, an improvement over the other methods. At $K = 1$, the margin widens sharply, as Agentic and Soft top-$K$ provide about a threefold reduction. Interestingly, a single view chosen by Agentic or Soft top-$K$ is more informative than two views placed by either of the fixed view methods.


What separates the methods is how gracefully their error increases as the view budget decreases (Fig.~\ref{fig:field}c). Between $K=90$ and $K=1$, the errors of Agentic and Soft top-$K$ increase by only $1.7\times$ and $1.8\times$, respectively, whereas the errors of Gumbel, Uniform, and Golden increase by $4.8\times$, $4.9\times$, and $5.0\times$, respectively. Thus, although Gumbel also learns the sampling mask, its error increases at a rate comparable to the fixed-view methods.
Agentic adds consistency to that accuracy. At $K = 2$ and $K = 1$ it has the
smallest standard deviation of any method 	($\pm 0.0065$ and $\pm 0.0043$, respectively), and at $K = 1$ its spread is very close to its spread at $K = 90$. Gumbel behaves in the opposite way, where it produces very high standard deviations at lower budgets. The rankings between the methods are summarized in Fig.~\ref{fig:field}(d).



\begin{figure}[htbp]
  \centering
  \includegraphics[width=\textwidth]{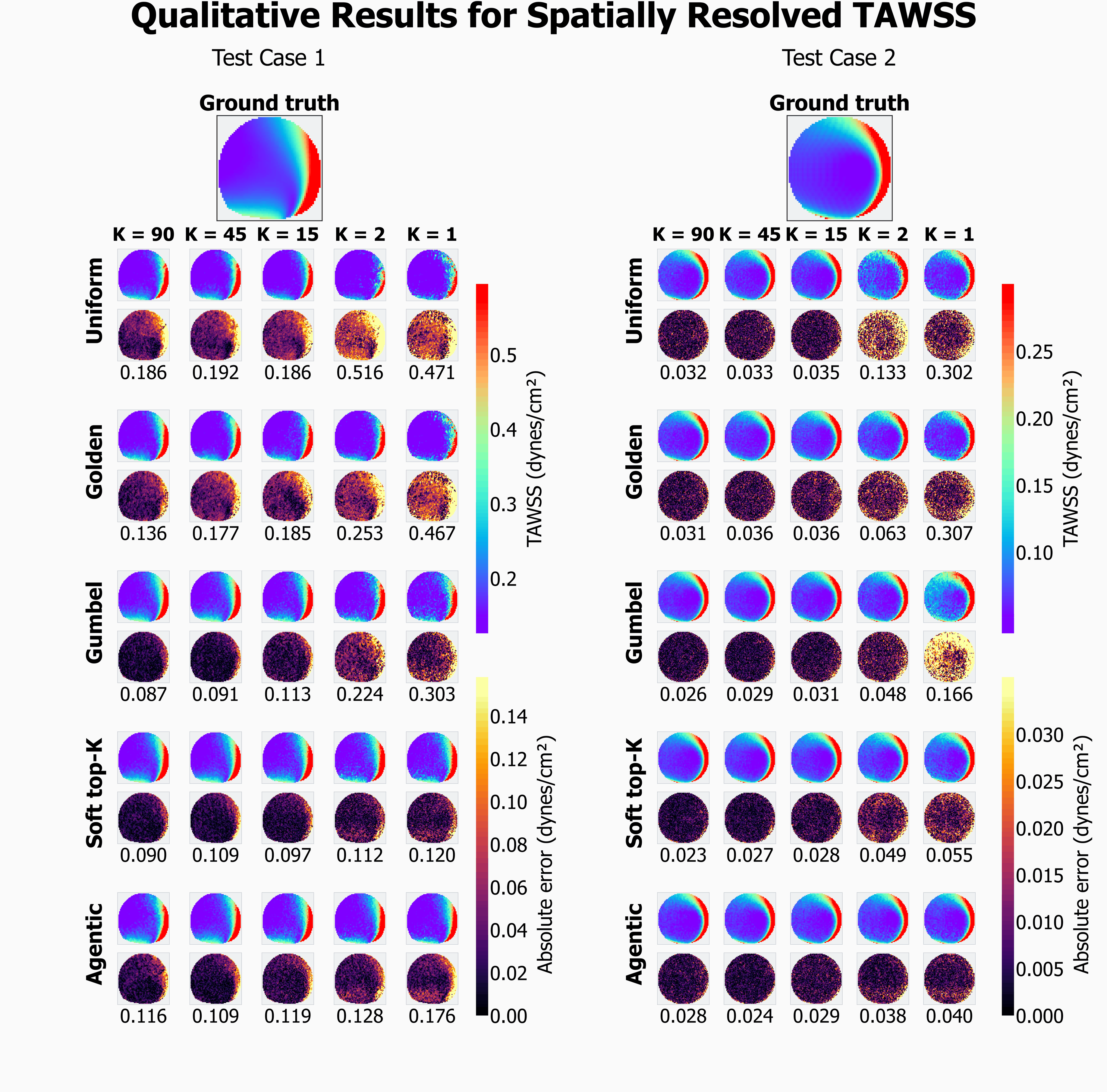}
  \caption{Predicted spatially resolved TAWSS for two test cases, shown for all
  five methods at view budgets $K$. Each panel gives the prediction with the
  Rel-$L_2$ error of that case below it, alongside the ground truth. The cases were selected at random from the test set
  and the per-case errors therefore differ from the run-averaged values in Table~\ref{tab:idealized-field-results}.}
  \label{fig:qualitative}
\end{figure}

Qualitatively, Fig.~\ref{fig:qualitative} shows that almost all methods
capture the general TAWSS pattern, and that the error comes mostly from a
magnitude offset and from noisy predictions. At $K = 2$ and $K = 1$, the
three methods with the highest error give visibly noisier maps, most clearly
for case 1, indicating that their masks retain less of the information needed
to resolve the field.

\begin{table}[ht]
\centering
\caption{Field prediction errors (mean $\pm$ standard deviation across five runs). 
\textbf{Bold} indicates the lowest mean error at each view budget. At $K=90$, Agentic and Soft top-$K$ have equal mean error to four decimal places and are both marked in bold. At $K=1$, Uniform and Golden use the same view, and their small differences arise from non-deterministic GPU arithmetic during training.}
\label{tab:idealized-field-results}
\setlength{\tabcolsep}{4pt}
\resizebox{\linewidth}{!}{%
\begin{tabular}{lccccc}
\toprule
Method & $K = 90$ & $K = 45$ & $K = 15$ & $K = 2$ & $K = 1$ \\
\midrule
Uniform  & $0.0391 \pm 0.0056$ & $0.0401 \pm 0.0052$ & $0.0432 \pm 0.0069$ & $0.1547 \pm 0.0217$ & $0.1925 \pm 0.0165$ \\
Golden   & $0.0386 \pm 0.0054$ & $0.0413 \pm 0.0061$ & $0.0439 \pm 0.0064$ & $0.0706 \pm 0.0090$ & $0.1924 \pm 0.0167$ \\
Gumbel   & $0.0345 \pm 0.0033$ & $0.0391 \pm 0.0073$ & $0.0416 \pm 0.0058$ & $0.0738 \pm 0.0289$ & $0.1660 \pm 0.0435$ \\
Soft top-$K$ & $\mathbf{0.0343 \pm 0.0045}$ & $0.0348 \pm 0.0036$ & $0.0407 \pm 0.0062$ & $0.0554 \pm 0.0122$ & $0.0623 \pm 0.0129$ \\
Agentic  & $\mathbf{0.0343 \pm 0.0052}$ & $\mathbf{0.0347 \pm 0.0063}$ & $\mathbf{0.0384 \pm 0.0051}$ & $\mathbf{0.0544 \pm 0.0065}$ & $\mathbf{0.0588 \pm 0.0043}$ \\
\bottomrule
\end{tabular}%
}
\end{table}

\subsection{Spatially Averaged TAWSS}
\label{sssec:spatially-averaged-tawss}

\begin{figure}[htbp]
  \centering
  \includegraphics[width=\textwidth]{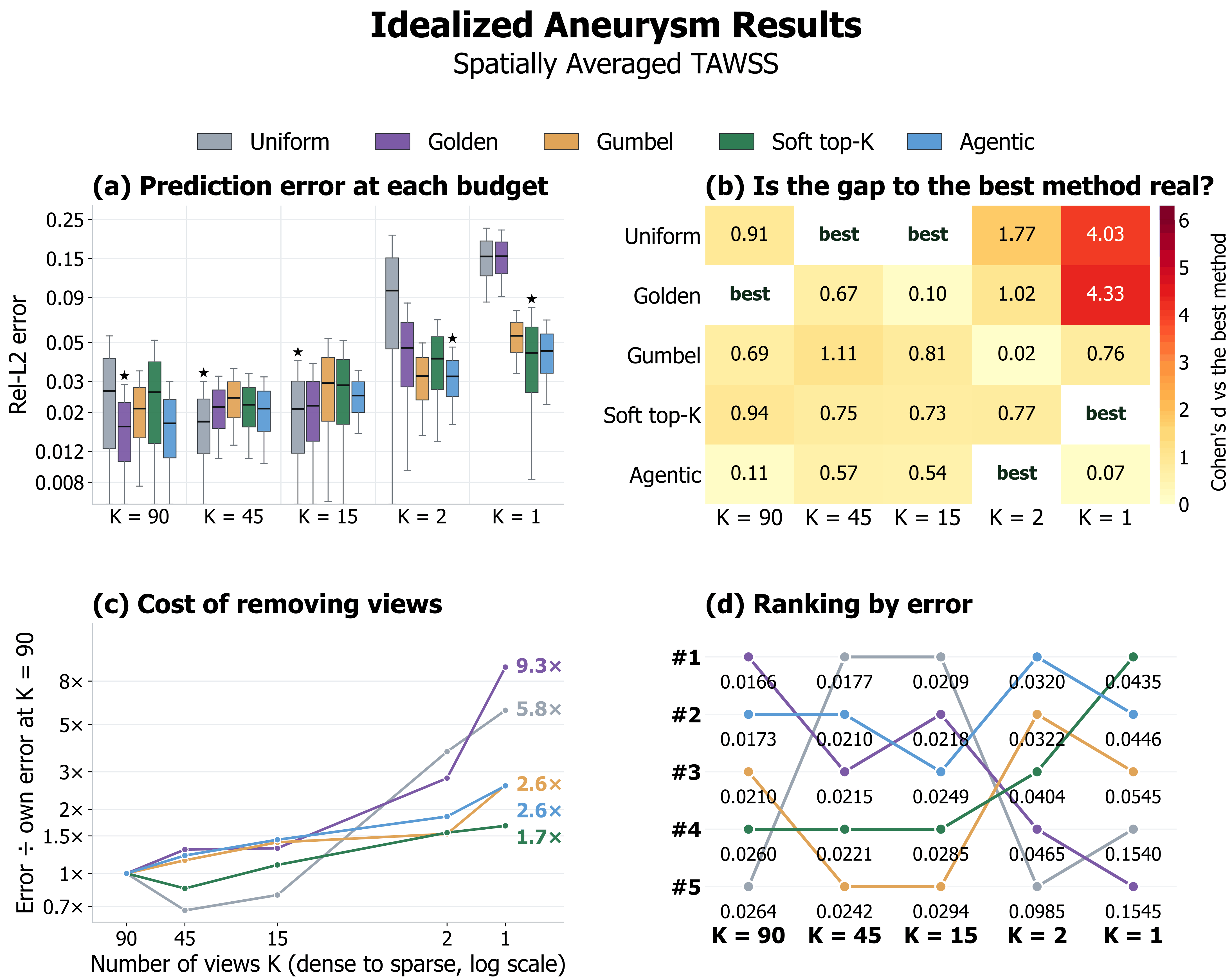}
  \caption{Spatially averaged TAWSS errors for view budgets $K$. Panels
  follow Fig.~\ref{fig:field}, with means and standard deviations computed
  over five runs. In panel (a), the error axis is
  logarithmic, so lower whiskers that fall below the lowest value shown,
  including those where the mean minus two standard deviations is negative,
  are truncated at the bottom of the axis.}
  \label{fig:scalar}
\end{figure}

Fig.~\ref{fig:scalar} and Table~\ref{tab:idealized-scalar-results}
show the results for the spatially averaged TAWSS. This target is easier to
predict than the field, and every method has a lower error here compared to the spatially resolved TAWSS at every budget. At $K \ge 15$, choosing the views does not help in obtaining consistent improvement (Fig.~\ref{fig:scalar}a). The
errors also do not always rise as the views are removed. For example, the Uniform method goes from $0.0264$ at $K = 90$ down to $0.0177$ at $K = 45$. The picture changes at $K \le 2$, where Agentic, Gumbel, and Soft top-$K$ obtain distinct improvements (Fig.~\ref{fig:scalar}b).


The effect of removing views is visualized in Fig.~\ref{fig:scalar}(c). From $K = 90$ to $K = 1$, the error grows by
$5.8\times$ for Uniform and $9.3\times$ for Golden, against $1.7\times$ for
Soft top-$K$ and about $2.6\times$ for Agentic and Gumbel. Golden's ratio is
large partly because its $K = 90$ error is low. The rankings (Fig.~\ref{fig:scalar}d) have no clear winner across all budgets. However, Agentic has the most consistent acceptable performance and it is the only method that never drops below third.  




\begin{table}[ht]
\centering
\caption{Scalar (spatially averaged) prediction errors (mean $\pm$ std over 5 runs). \textbf{Bold} marks the best value in each column. At $K=1$, Uniform and Golden use the same view, and their small differences arise from non-deterministic GPU arithmetic during training.}
\label{tab:idealized-scalar-results}
\setlength{\tabcolsep}{4pt}
\resizebox{\linewidth}{!}{%
\begin{tabular}{lccccc}
\toprule
Method & $K = 90$ & $K = 45$ & $K = 15$ & $K = 2$ & $K = 1$ \\
\midrule
Uniform  & $0.0264 \pm 0.0140$ & $\mathbf{0.0177 \pm 0.0061}$ & $\mathbf{0.0209 \pm 0.0092}$ & $0.0985 \pm 0.0526$ & $0.1540 \pm 0.0346$ \\
Golden   & $\mathbf{0.0166 \pm 0.0061}$ & $0.0215 \pm 0.0053$ & $0.0218 \pm 0.0081$ & $0.0465 \pm 0.0186$ & $0.1545 \pm 0.0317$ \\
Gumbel   & $0.0210 \pm 0.0067$ & $0.0242 \pm 0.0056$ & $0.0294 \pm 0.0116$ & $0.0322 \pm 0.0087$ & $0.0545 \pm 0.0106$ \\
Soft top-$K$ & $0.0260 \pm 0.0127$ & $0.0221 \pm 0.0056$ & $0.0285 \pm 0.0114$ & $0.0404 \pm 0.0134$ & $\mathbf{0.0435 \pm 0.0176}$ \\
Agentic  & $0.0173 \pm 0.0063$ & $0.0210 \pm 0.0054$ & $0.0249 \pm 0.0049$ & $\mathbf{0.0320 \pm 0.0075}$ & $0.0446 \pm 0.0112$ \\
\bottomrule
\end{tabular}%
}
\end{table}

\subsection{Interpretation of Selected Views}
\label{ssec:view_interpretation}

\begin{figure}[htbp]
  \centering
  \includegraphics[width=\textwidth]{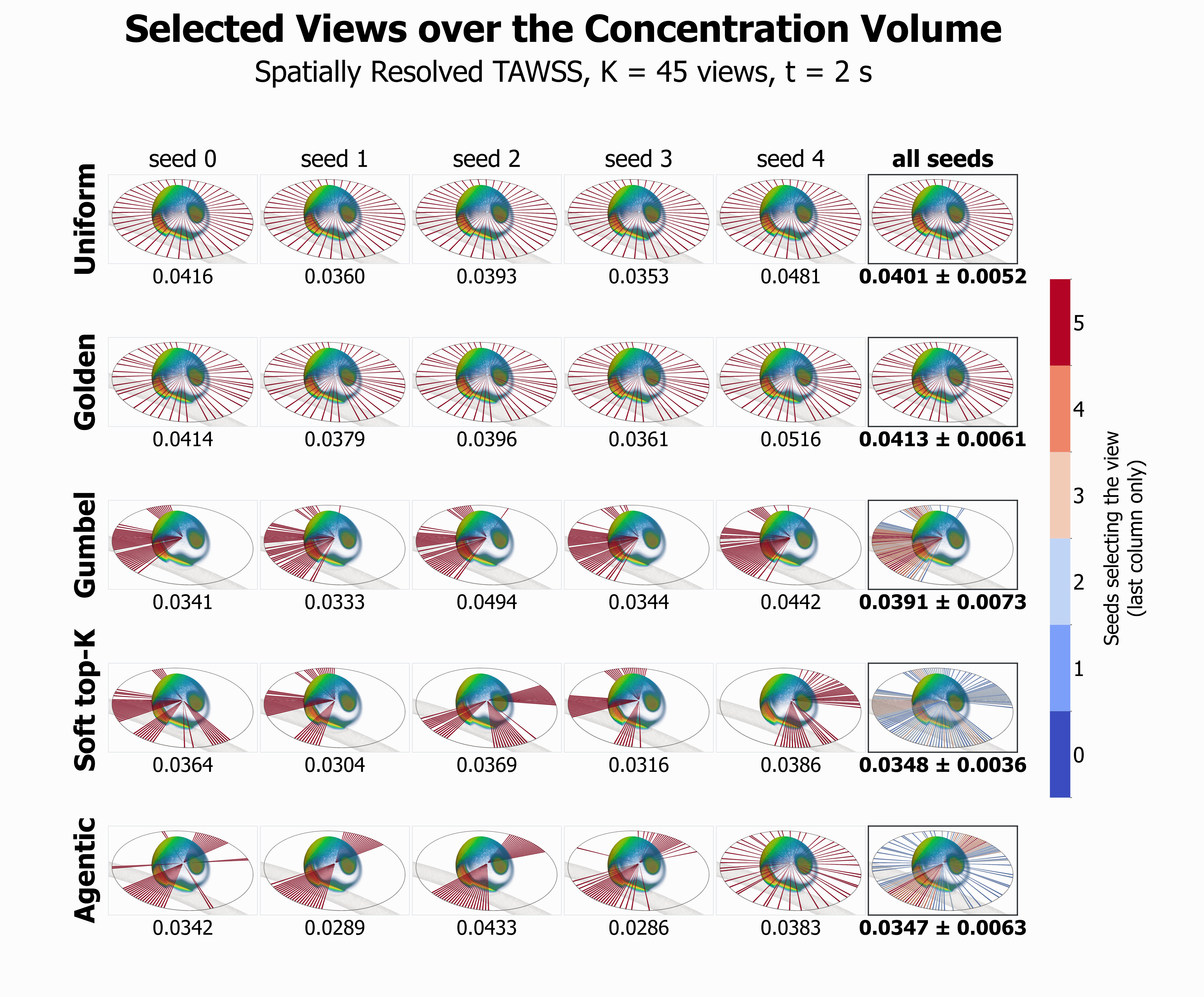}
  \caption{Views selected by each method for the spatially resolved target at
  $K = 45$, drawn over the contrast concentration volume at $t = 2$~s.
  Each ray marks one acquired projection angle and the outline shows the full
  $360^{\circ}$ candidate ring of $N_v = 180$ views spaced $2^{\circ}$ apart.
  The first five columns are the five runs, each labeled with its test
  Rel-$L_2$. The last column pools the runs and is labeled with the mean
  $\pm$ standard deviation over them. In this last column, the rays are colored by how many of the five runs selected each view (color scale at the right).}
  \label{fig:views_field_K45}
\end{figure}

\begin{figure}[htbp]
  \centering
  \includegraphics[width=\textwidth]{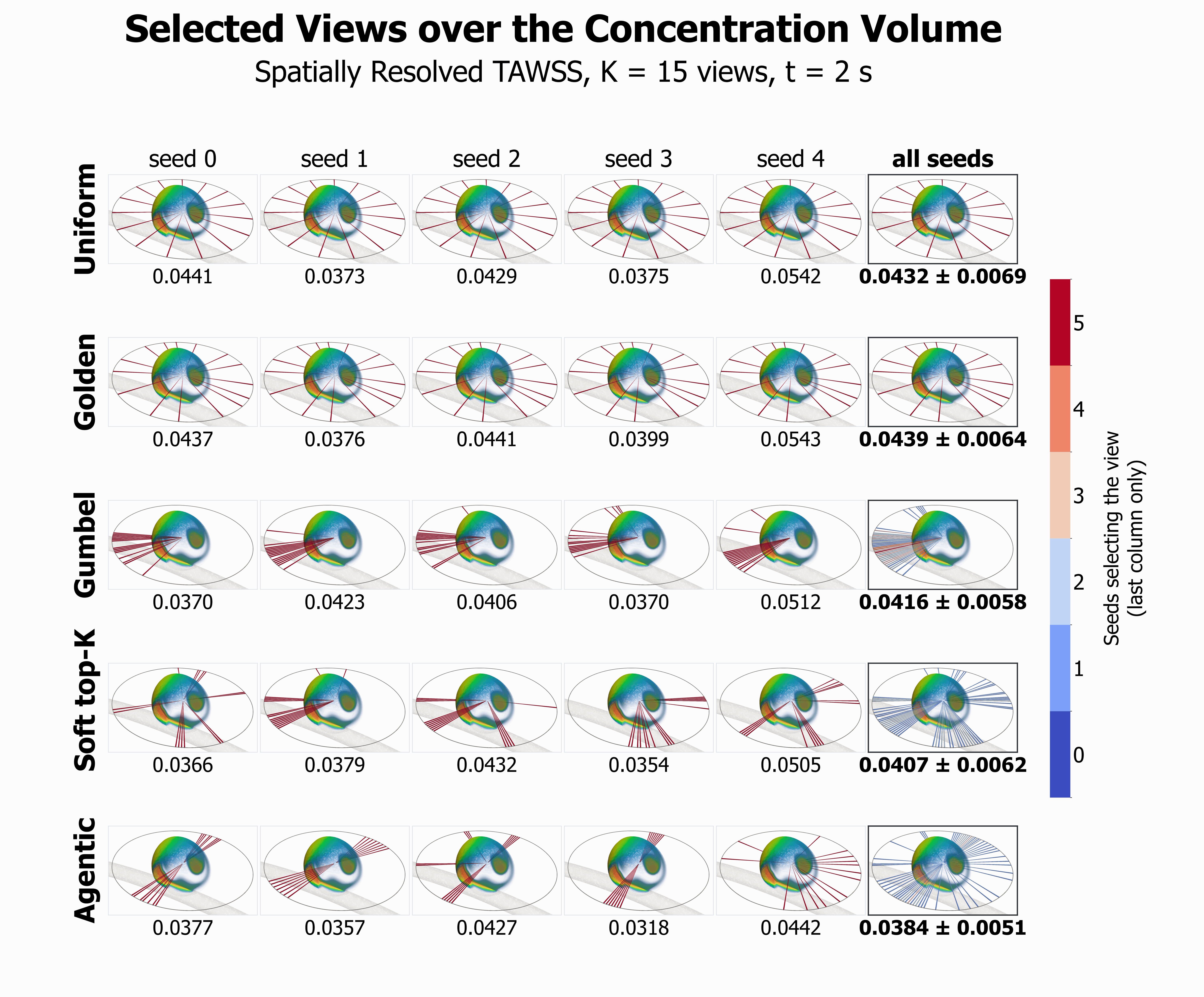}
  \caption{Views selected by each method for the spatially resolved target at
  $K = 15$, drawn over the contrast concentration volume at $t = 2$~s.
  Layout as in Fig.~\ref{fig:views_field_K45}.}
  \label{fig:views_field_K15}
\end{figure}

\begin{figure}[htbp]
  \centering
  \includegraphics[width=\textwidth]{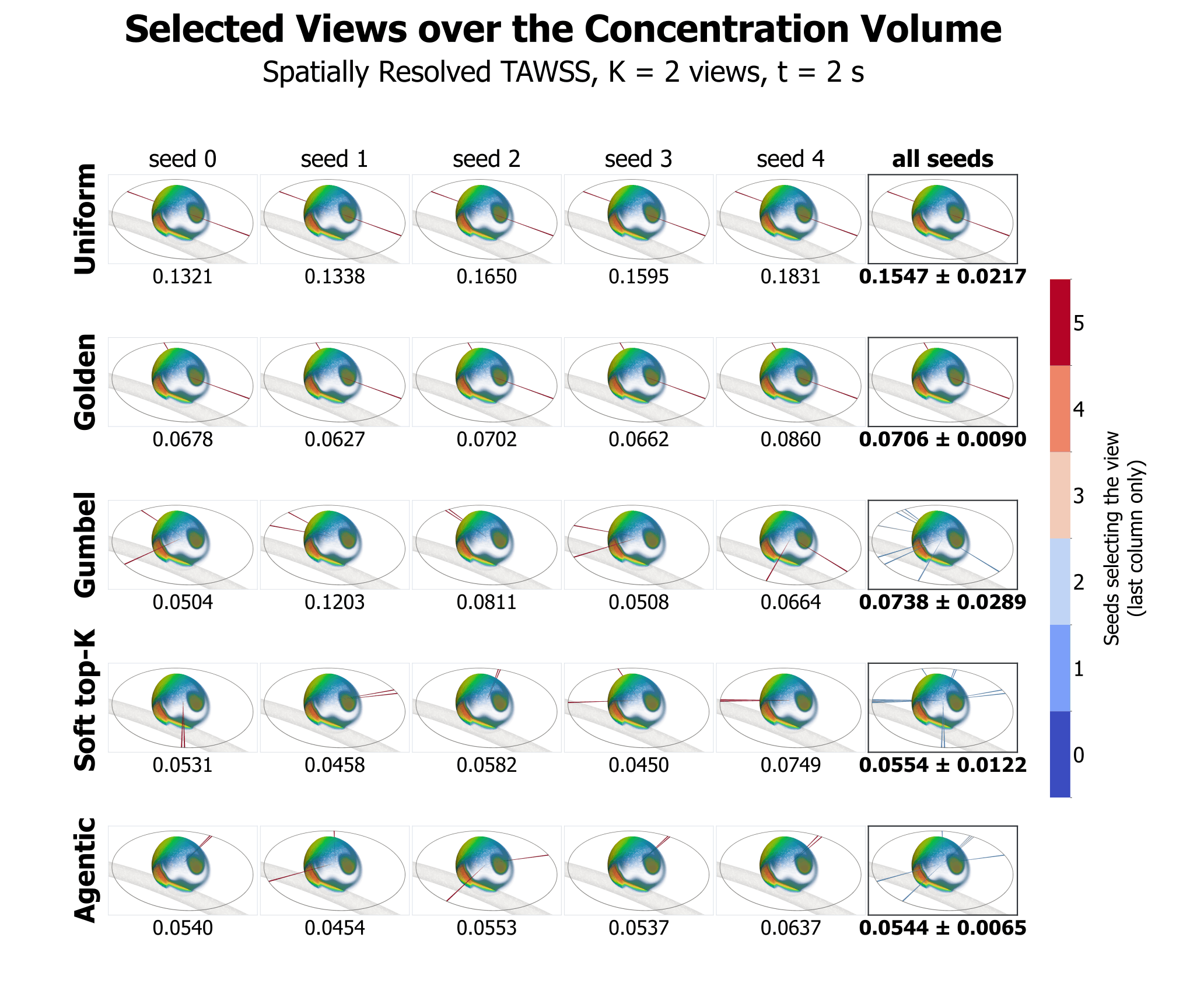}
  \caption{Views selected by each method for the spatially resolved target at
  $K = 2$, drawn over the contrast concentration volume at $t = 2$~s.
  Layout as in Fig.~\ref{fig:views_field_K45}.}
  \label{fig:views_field_K02}
\end{figure}

To understand the behavior of the selected masks, we examined and visualized
the selected angles for the spatially resolved TAWSS. Figs.~\ref{fig:views_field_K45}--\ref{fig:views_field_K02}
show the selected views for each method at three budgets, $K \in \{45, 15,
2\}$, drawn over a volumetric rendering of the concentration in which high
concentration is opaque and low values are faded.

At $K = 45$ (Fig.~\ref{fig:views_field_K45}), the selected views of all three
learned methods, regardless of method and seed, tend to lie in the region
between the high-concentration region produced by the impingement at the
frontal part of the aneurysm sac and the low-concentration region at its
sides. This region has two distinguishing features. First, it is where the
concentration gradient is steepest, and the gradient is where the information
lies for similar transport problems~\cite{sharma2019analytic,elhadidy2026wall}. Second, it is where the different aneurysm cases differ most from one another, since the size of the high-concentration impingement region is set by the parameters varied in this dataset~\cite{elhadidy2026sle}, which in turn control the extent of the high- and low-concentration regions.

Comparing the seeds at $K = 45$ (Fig.~\ref{fig:views_field_K45}, last
column), Gumbel is the most consistent method although the least accurate among the three learned methods.
Agentic is noticeably more consistent than Soft top-$K$, and Soft top-$K$ is the
least consistent (most sensitive to the seed). The patterns differ as
well. Gumbel groups its views into two clusters, one slightly larger than the
other, giving complementary coverage of the angular range. Soft top-$K$ is the opposite where almost no view appears in all five seeds, but every seed follows the same pattern of three or four contiguous clusters. Agentic behaves differently again, giving mirrored or symmetric arrangements in the first four seeds, while the fifth resembles a perturbed or locally refined golden-angle mask. 


At $K = 15$ (Fig.~\ref{fig:views_field_K15}), the same behavior continues,
except that Agentic becomes less consistent than at $K = 45$. Gumbel still
clusters its views, but now into a single region at the band between the high- and low-concentration regions. Soft top-$K$ keeps almost exactly the same
behavior, clustering into three or four groups, and Agentic retains its
symmetric arrangements with the best accuracy at this budget.

At $K = 2$ (Fig.~\ref{fig:views_field_K02}), the most noticeable behavior is
that several seeds place their two views adjacent to each other, indicating a
tendency to take closely spaced readings of the same region. These observations suggest there is no single best set of views for this task.
Very different masks reach similar accuracy, and the most repeatable method in this case is interestingly the least accurate one. The same non-uniqueness has been reported for optimal sensor placement, where several different configurations carry comparable information~\cite{ercan2021optimal}, and we discuss this in the Discussion Section.

\subsection{Agreement Between Masks Across Tasks and Budgets}
\label{ssec:task_budget_masks}
\begin{figure}[htbp]
  \centering  \includegraphics[width=\textwidth]{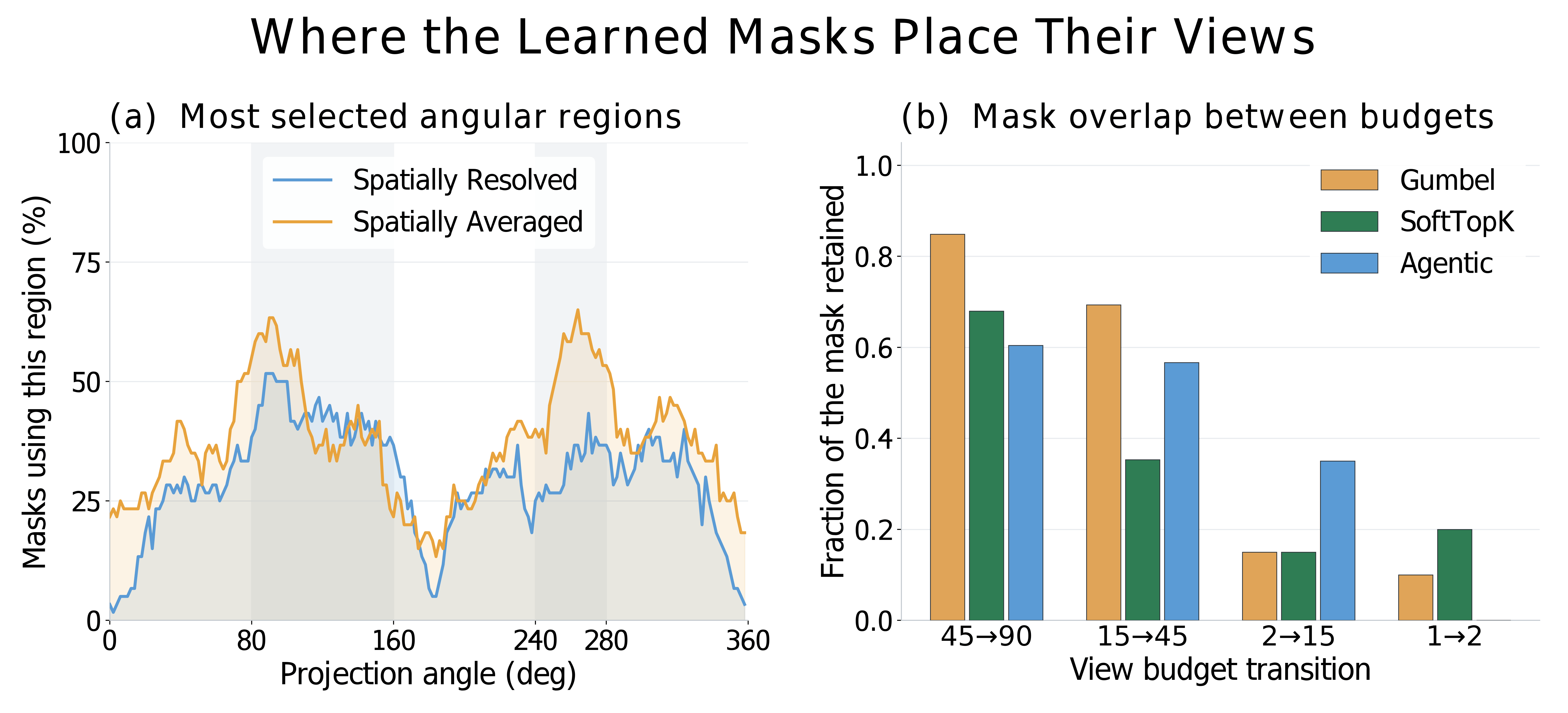}
  \caption{Angular structure of the learned masks for Gumbel, Soft top-$K$ and
  Agentic with five runs per target and budget (60 masks per target at
  $K \leq 45$).
  \textbf{(a)} Percentage of masks that place at least one view within
  $\pm 5^{\circ}$ of each angle, for $K \leq 45$. The angle axis spans the
  full $360^{\circ}$ gantry rotation. Shaded bands mark
  $80^{\circ}$--$160^{\circ}$ and $240^{\circ}$--$280^{\circ}$.
  \textbf{(b)} Fraction of a mask that reappears at the next higher budget,
  averaged over runs and both targets.}
  \label{fig:views_angles}
\end{figure}

To better understand the agreement and distinction between the learned masks, we measure the differences between the two targets at all budgets, and between budgets for the same method in Fig.~\ref{fig:views_angles}. For agreement between targets, we count how often each angle is used. We
pooled the masks from the three learned methods, the five runs, and all
budgets $K \leq 45$, giving 60 masks per target. We count a mask
as using an angle if it has a view within $\pm 5^{\circ}$ of it, that is, we count whether the two targets agree on angular regions rather than on precise individual views~\cite{sechidis2019stability,hamer2021robust}.

The two targets agree in part as shown in Fig.~\ref{fig:views_angles}a. Both use the band from $80^{\circ}$ to $160^{\circ}$ heavily and the least used regions, around $0^{\circ}$--$20^{\circ}$, $170^{\circ}$--$190^{\circ}$ and $350^{\circ}$--$360^{\circ}$, are those aligned with the main arterial flow direction.


For agreement between budgets, we measure how much of a mask survives when
more views become available. For two budgets $K_1 < K_2$, we take the
fraction of the $K_1$ views that also appear in the $K_2$ mask of the same
method. A high fraction means the sparse mask is close to a subset of the
denser one. Agreement is high between the dense budgets
(Fig.~\ref{fig:views_angles}b): $0.60$--$0.85$ from $K = 45$ to $K = 90$,
with Gumbel the highest and Agentic the lowest, and $0.35$--$0.69$ from
$K = 15$ to $K = 45$, where Gumbel is still the highest. It then collapses to $0.15$--$0.35$ from $K = 2$ to $K = 15$, where Agentic is almost double the other methods.
 
\subsection{Agentic AI Behavior and Efficiency}
\label{ssec:agentic_analysis}
We used the agentic search logs to track how the proposed view sets evolved
and how each change affected validation performance. For every trial across
the 50 runs, 390 in total, we recorded the candidate mask, its validation
error, and the reasoning the agent gave for its next proposal. Because the
runs are independent for each task and budget, we could ask whether the
agents follow any common pattern and whether each agentic proposal builds on earlier trials in the same run, on a separate analysis, or on neither. We also
examined which design concepts appear, when in a run they are used, and how
often one concept follows another. We grouped the reasoning into 17
categories, placing related ideas together, such as uniform and grid
placements, or block and contiguous bands. Table~\ref{tab:concepts} defines
these reasoning concepts. Each trial was assigned to categories by keyword
matching on its stated candidate construction, so a single trial can belong
to more than one category. For example, a trial described as a uniform
baseline counts as both uniform placement and a baseline/control. Counts and
percentages of concepts therefore need not sum to the number of trials or to
100\%. Alongside the reasoning, we tracked the validation error at each trial and whether it falls as the search proceeds, and we examined the resulting agentic search trajectories.

As a human validation step and to check the logs, we reran selected trials manually using the recorded mask and epoch count, and compared the resulting validation error with the value reported in the log.

\begin{figure}[htbp]
  \centering
  \includegraphics[width=\textwidth]{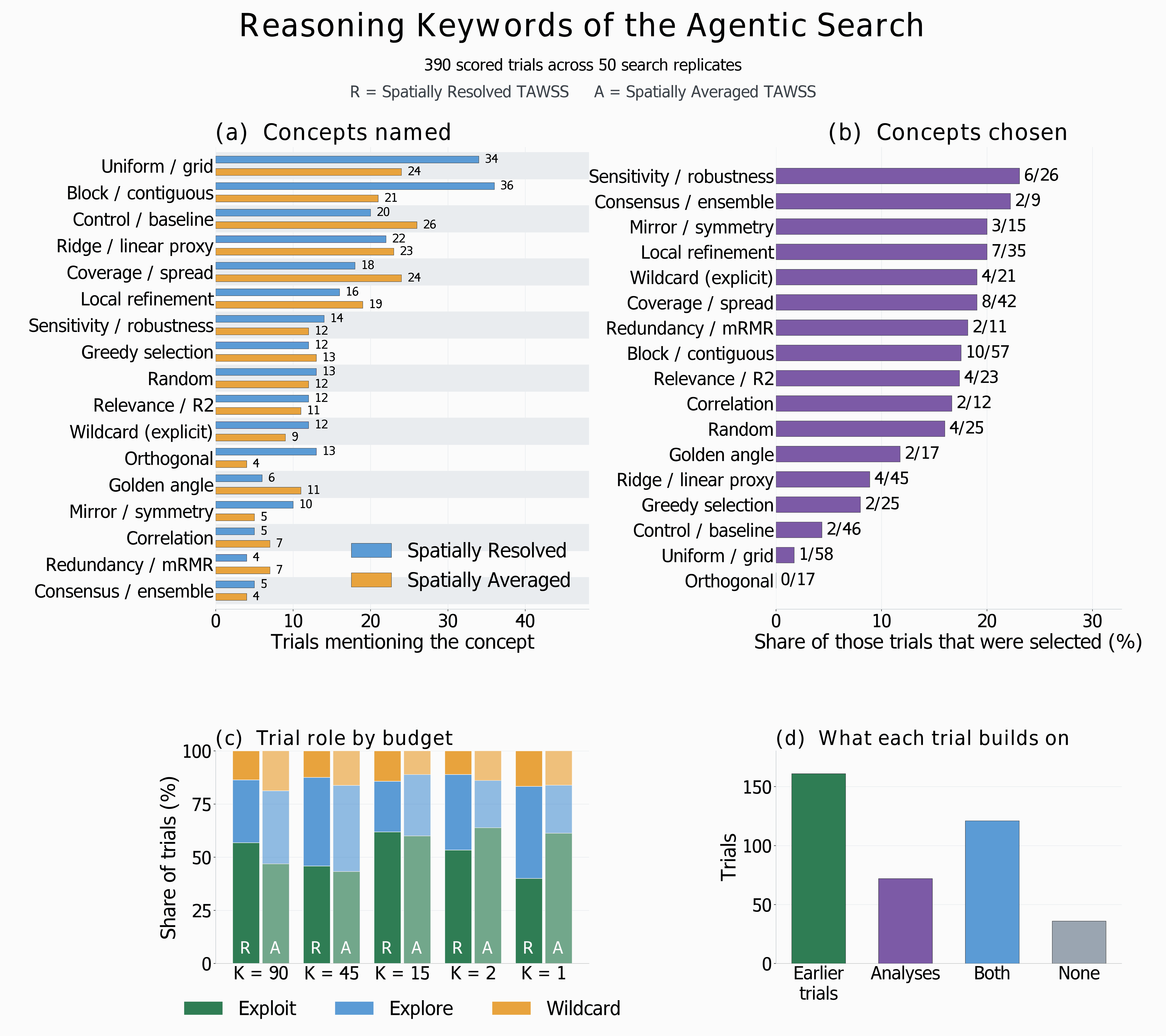}
  \caption{Reasoning of the Agentic search over all 390 scored trials from 50
  runs. Design concepts are coded by keyword match on the stated
  candidate construction and a trial may name more than one concept. Each concept is
  defined in Table~\ref{tab:concepts}.
  \textbf{(a)} Trials naming each concept. 
  \textbf{(b)} For each concept, the ratio of the trials that were
  finally chosen to the trials that proposed it.
  \textbf{(c)} Share of trials by role at each budget, for the spatially
  resolved (R) and spatially averaged (A) targets. Roles are the agent's own
  labels: \emph{exploit} refines a construction that already worked in that
  run, \emph{explore} tests an unproven construction or a deliberate control,
  and \emph{wildcard} is an unconventional idea outside the current line of
  reasoning, which the protocol required each run to attempt.
  \textbf{(d)} The predecessors each trial declares: $41\%$ build only on
  earlier trials in the same run, $31\%$ on both earlier trials and separate
  analyses, $18\%$ on analyses alone, and $9\%$ declare none.}
  \label{fig:reasoning}
\end{figure}

\begin{table*}[htbp]
  \centering
  \caption{The design concepts coded in Fig.~\ref{fig:reasoning}. A trial is
  assigned a concept when its stated candidate construction contains one of
  the listed keywords, so a single trial may carry several concepts.
  \emph{Proposed} is the number of trials naming the concept and
  \emph{chosen} the number of those that became a run's final used mask.}
  \label{tab:concepts}
  \small
  \begin{tabular}{@{}l r r p{0.53\textwidth}@{}}
    \toprule
    Concept & Proposed & Chosen & What it means \\
    \midrule
    Uniform / grid & 58 & 1 &
      Views placed at equal angular spacing, optionally rotated by an offset.
      The standard sparse-view construction, used here mostly as a starting
      point. \\
    Block / contiguous & 57 & 10 &
      Views drawn from one or more contiguous angular bands rather than spread
      over the full range, including bands that wrap past $360^{\circ}$. \\
    Control / baseline & 46 & 2 &
      A candidate proposed to test a claim rather than to win, such as a
      deliberately poor placement expected to fail. \\
    Ridge / linear proxy & 45 & 4 &
      Views ranked by a cheap supervised surrogate, typically a ridge or
      least-squares fit from single views to the target. \\
    Coverage / spread & 42 & 8 &
      Placement driven by an explicit geometric spacing criterion such as
      farthest-point, stratified or low-discrepancy sampling. \\
    Local refinement & 35 & 7 &
      Small perturbations of an existing mask: shifting a view by a few
      degrees, jittering, or swapping in an angular neighbor. \\
    Sensitivity / robustness & 26 & 6 &
      Candidates built or screened for stability, checking that performance
      survives a small change of views or of seed. \\
    Greedy selection & 25 & 2 &
      Views added one at a time, each chosen to maximize an incremental
      criterion given those already selected. \\
    Random & 25 & 4 &
      Views drawn at random, used almost entirely as a reference against
      which structured constructions are judged. \\
    Relevance / $R^{2}$ & 23 & 4 &
      Views weighted by an estimated per-view importance, for example the
      variance explained in the target. \\
    Wildcard (explicit) & 21 & 4 &
      A deliberately unconventional idea outside the current line of
      reasoning. \\
    Orthogonal & 17 & 0 &
      Pairs or sets of mutually perpendicular views, the classical
      minimal-redundancy choice. Never selected here. \\
    Golden angle & 17 & 2 &
      Successive views separated by $137.5^{\circ}$, giving incoherent,
      quasi-uniform coverage for any number of views. \\
    Mirror / symmetry & 15 & 3 &
      Views chosen in symmetric pairs about an anatomical or geometric axis,
      or placed to probe an asymmetry. \\
    Correlation & 12 & 2 &
      Views selected or discarded using the correlation between the
      projections they produce. \\
    Redundancy / mRMR & 11 & 2 &
      Explicitly penalizing views that duplicate information already
      captured, as in minimum-redundancy maximum-relevance selection. \\
    Consensus / ensemble & 9 & 2 &
      A mask assembled by combining the results of several earlier trials or
      several settings of a hyperparameter. \\
    \bottomrule
  \end{tabular}
\end{table*}

\begin{figure}[htbp]
  \centering

  \includegraphics[width=0.95\linewidth]{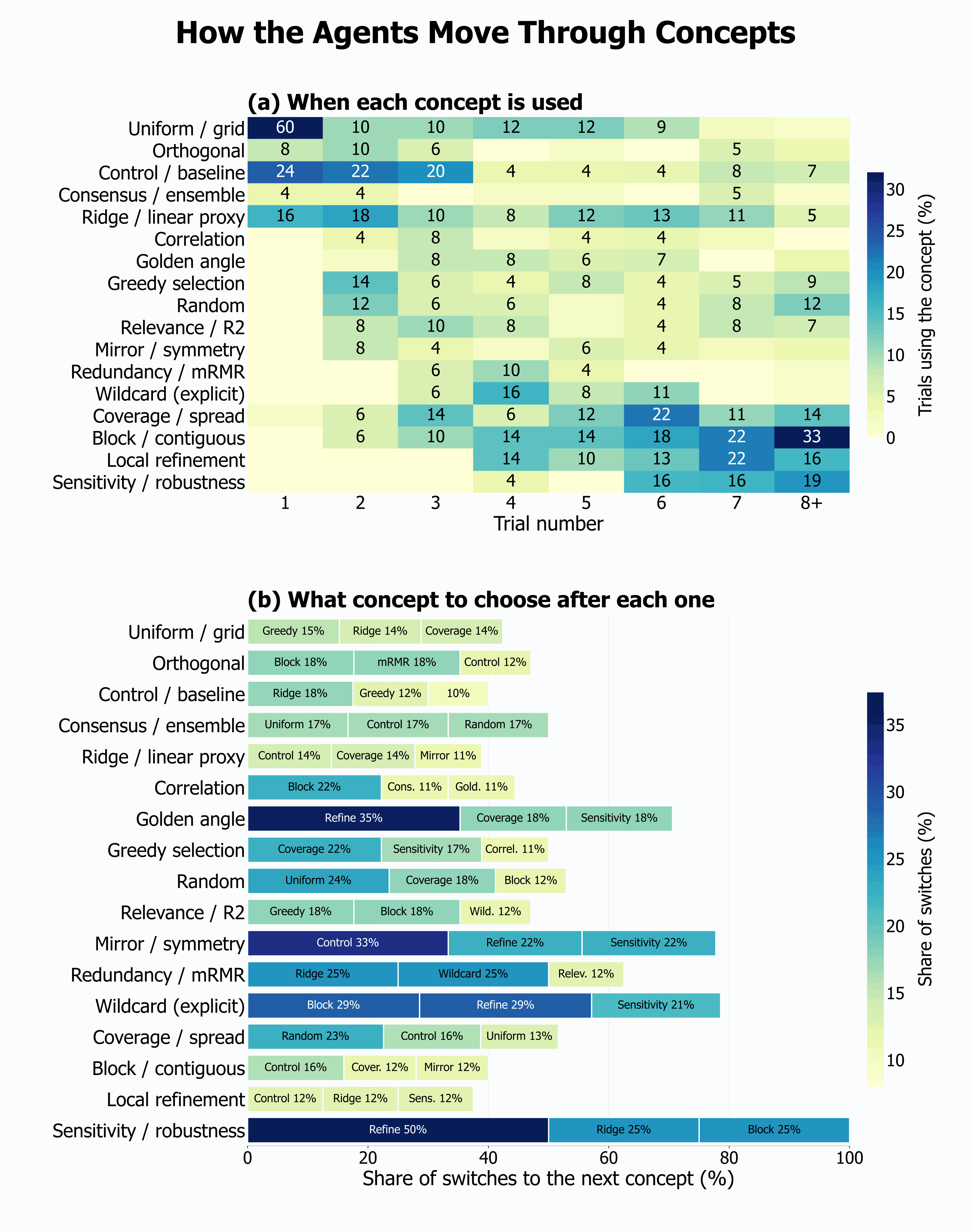}
  \caption{How the agents move through design concepts, using the same
  keyword coding as Fig.~\ref{fig:reasoning} over 390 trials from 50
  independent searches. Concepts are ordered by their typical position in a
  run, and both panels share that order.
  \textbf{(a)} Percentage of trials using each concept at each trial number,
  with the eighth and later trials pooled. Uniform sampling dominates the
  first trial, while local refinement and sensitivity checks appear only
  from the fourth trial onward.
  \textbf{(b)} For each concept on the left, the three concepts the agent
  most often chooses in the next trial, ordered from most to least common and colored by that share. A switch is a concept used in the next trial but not in the
  current one.}
  \label{fig:strategy_sequence}
\end{figure}

\subsubsection{Reasoning Concepts and Sequential Reasoning}
\label{sssec:reasoning_concepts}

Table~\ref{tab:concepts} lists the design concepts the agent used. Fig.~\ref{fig:reasoning} summarizes how these concepts were used.
Fig.~\ref{fig:reasoning}(a) gives the number of trials naming each concept,
separately for the spatially resolved and spatially averaged targets.
Proposing a concept does not mean it produces the best validation error, therefore, Fig.~\ref{fig:reasoning}(b) reports, for each concept, the ratio of the
trials that were finally selected to the trials that proposed it.
Fig.~\ref{fig:reasoning}(c) shows how the agent behaved at each budget,
whether it refined a construction that had already worked in that run
(exploit), tested an unproven one (explore), or tried an idea outside its
current line of reasoning (wildcard). Fig.~\ref{fig:reasoning}(d) shows what
each trial declared it was built on, that is, earlier trials only, a separate analysis only, both, or not mentioned.

Next, we looked at the order in which the concepts appear. Fig.~\ref{fig:strategy_sequence}(a) shows where in a run each concept tends
to be proposed or selected, and Fig.~\ref{fig:strategy_sequence}(b) gives the
transition frequencies between concepts, that is, which concept tends to
follow which, with the three most likely successors and their probabilities. The most frequently proposed concept was uniform or grid placement, which is
equivalent to the uniform baseline. It appears in 58 trials and opens 60\% of the 50 runs (Fig.~\ref{fig:strategy_sequence}a), yet it produced the best validation error only
once. After testing it, the agent most often moved on to greedy selection, a
ridge-based analysis, or an explicit coverage criterion, each following the uniform placement in about 15\% of switches
(Fig.~\ref{fig:strategy_sequence}b). Uniform placement therefore serves as a
starting point rather than as a candidate solution. Block or contiguous placement was proposed 57 times and selected as the final
mask in 10 of them, the highest count of any concept and a selection ratio of
about 18\%. It was used more for the spatially resolved target, in 36 of the
57 trials. In sequence it appears from the second trial onward, most densely at trial 8 and beyond, and its most frequent successor category was a control trial (16\% of switches; Fig.~\ref{fig:strategy_sequence}b).

The highest selection-to-proposal ratio belonged to sensitivity or robustness
screening, at 6 of 26 trials. Candidates in this category were
built or tested for stability, checking that performance survives a small
change of views. In the sequence heat map, these trials never appear before trial 4 and are most common from trial 6 onward, suggesting they serve as a
stability check on a candidate that had already performed well. They are most often followed by local refinement (50\% of switches), a small shift or swap of individual
views, which is consistent with adjusting a mask after seeing how sensitive it is (Fig.~\ref{fig:strategy_sequence}b).

Orthogonal placement, in which views are chosen in pairs separated by \(90^\circ\), was proposed 17 times but never selected as a final mask. It appeared mostly in the first three trials of a run, which shows it was an early guess that the agents tried and then dropped. This is notable because perpendicular views are the classical way to avoid redundancy. For predicting TAWSS, however, this choice did not help, since the best masks often placed views close together in the region of steep concentration gradient rather than spreading them apart. Its complete absence from the final masks is an example of a reasonable prior idea of the agent that the data did not support.

The balance between exploitation, exploration and wildcard trials is similar across budgets, at roughly 55\%, 30\% and 15\% respectively. The one visible
trend is for the spatially averaged target, where the share of exploitation
decreases as the budget grows (Fig.~\ref{fig:reasoning}c). As for what each
proposal was built on, most trials follow earlier ones, where 41\% declare only
previous trials, 31\% both previous trials and a separate analysis, 18\%
analyses alone, and 9\% declare nothing (Fig.~\ref{fig:reasoning}d). 


\subsubsection{Agent Search Trajectories}
\label{sssec:search_trajectories}

\begin{figure}[htbp]
  \centering
  \includegraphics[width=\textwidth]{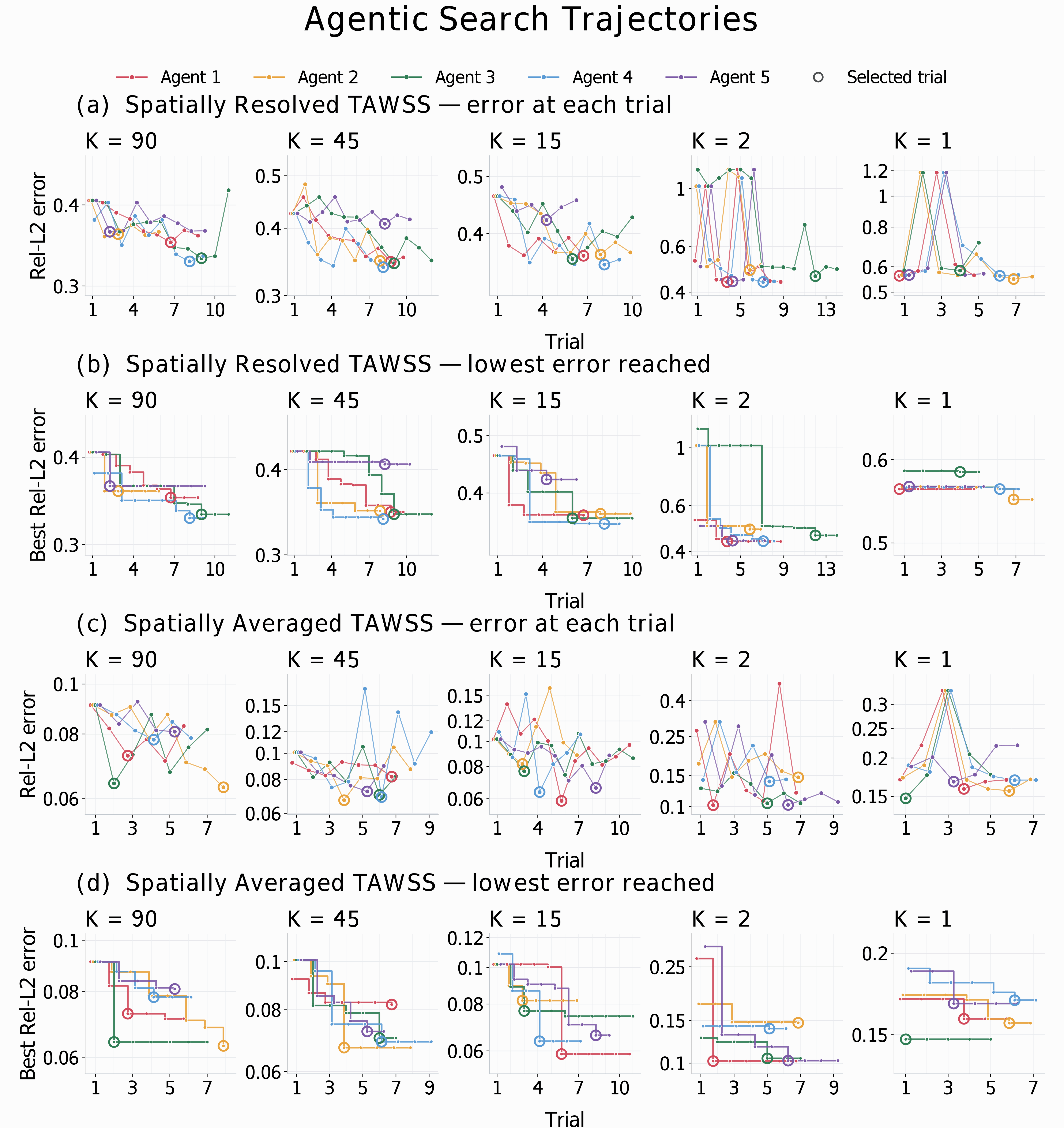}
  \caption{Search trajectories of the Agentic view selector: five independent
  search agents per panel, one column per view budget. \textbf{(a)} Validation
  error of the candidate proposed at each trial for spatially resolved TAWSS.
  \textbf{(b)} The lowest error reached up to that trial, i.e.,\ the cumulative
  minimum of panel (a). \textbf{(c)}, \textbf{(d)} The same two views for spatially
  averaged TAWSS. Rings mark the trial each agent finally selected.}
  \label{fig:search}
\end{figure}

Fig.~\ref{fig:search} shows the search trajectory of each of the five independent agents at every view budget. Each point in Fig.~\ref{fig:search}(a,c) corresponds to one candidate projection subset proposed and evaluated during the search. Since consecutive trials represent different discrete masks, rather than incremental updates of the same mask, the validation error is not expected to decrease monotonically. A trial can therefore perform worse than the current best candidate even if the overall search later identifies a lower error subset. The rings in the figure indicate the trial corresponding to the subset ultimately selected by each agent.

The cumulative minimum curves in Fig.~\ref{fig:search}(b,d) provide a complementary view of the same searches. At each trial, each curve reports the lowest validation error encountered from the beginning of the search through that specific trial. A decrease in these curves therefore indicates that the agent has identified a new candidate that improves on all of its previous proposals, while a flat segment indicates that subsequent candidates did not improve the best error found so far. 

For spatially resolved TAWSS, the 25 independent searches required between 5 and 14 trials, with an average of 8.36 candidate evaluations per search. The candidate that was ultimately selected was first evaluated, on average, at trial 6.24. Thus, the selected subset was typically identified before the search terminated, with an average of 2.12 additional candidate subsets evaluated afterward. The search length also varied across view budgets. The $K=1$ searches were the shortest, averaging 6.0 trials, whereas the $K=2$ searches averaged 9.0 trials and ranged from 7 to 14 evaluations. For $K=45$, the searches averaged 9.6 trials, and the selected candidate was first encountered relatively late, at trial 8.4 on average.

The spatially averaged TAWSS searches were somewhat shorter. Across the 25 runs, each agent evaluated between 5 and 11 candidate subsets, with an average of 7.24 trials. The selected candidate was first evaluated, on average, at trial 4.76, leaving approximately 2.48 additional evaluations before termination. The longest searches occurred at $K=15$, with an average of 9.0 trials, while the $K=1$ and $K=90$ searches averaged 6.2 and 6.4 trials, respectively.

The cumulative minimum curves also show that most searches improved through several intermediate candidates rather than identifying their best candidate immediately. For spatially resolved TAWSS, a new best validation error was found an average of 3.04 times after the initial trial. The corresponding average for spatially averaged TAWSS was 2.48 improvements per search. The number and timing of these improvements differed between agents even at the same budget $K$ as expected since the five runs at each setting are independent.


\subsubsection{Agentic AI Reproducibility: Human Validation}
\label{sssec:agentic_reproducibility}
The agent selects views by training each candidate subset and reading back
its validation error. These logged errors must be reproducible for the
search to be meaningful. To check this, we reran two individual search
trials directly from the agent logs. These are trials the agent evaluated
during its search (not the final models). The
two trials were chosen in advance to cover both targets and two view budgets:
a two-view field trial ($268^\circ$, $270^\circ$) and a 15-view scalar trial
($8^\circ$, $40^\circ$, $88^\circ$, $92^\circ$, $96^\circ$, $120^\circ$,
$160^\circ$, $188^\circ$, $200^\circ$, $240^\circ$, $266^\circ$, $270^\circ$,
$274^\circ$, $320^\circ$, $352^\circ$). Each was retrained from scratch with
the same settings the agent used during its search. The reruns reproduced the logged validation Rel-$L_2$ errors to within $0.02\%$ for the field trial and $3.95\%$ for the scalar trial. 


\subsection{Computational Cost}
\label{ssec:computational_cost}

Table~\ref{tab:computational_cost} compares the computational cost of the
view selection methods for the spatially resolved TAWSS at $K=45$. All timings
report compute time and were measured on an NVIDIA RTX 2080 Ti (11.3~GB). All
methods use the same predictor architecture with 4,202,978 trainable
parameters.

\begin{table}[t]
\centering
\caption{Computational cost of the view selection methods for the spatially
resolved TAWSS at $K=45$.}
\label{tab:computational_cost}
\resizebox{\columnwidth}{!}{%
\begin{tabular}{lcccc}
\toprule
\textbf{Method} &
\textbf{Selection (s)} &
\textbf{Predictor training (s)} &
\textbf{Total (min)} &
\textbf{Peak GPU memory (MB)} \\
\midrule
Uniform  & 0        & 892.5  & 14.9 & 561  \\
Golden   & 0        & 892.5  & 14.9 & 561  \\
Gumbel   & In-train & 895.7  & 14.9 & 562  \\
Soft top-$K$ & In-train & 1005.3 & 16.8 & 1053 \\
Agentic  & 570$^{\dagger}$ & 892.5  & 24.4 & 617$^{*}$ \\
\bottomrule
\end{tabular}%
}
\vspace{1mm}
\begin{flushleft}
\footnotesize
$^{\dagger}$Average of 9.6 search trials at $K=45$, each a 200-epoch
training run of 59.4~s.
$^{*}$Peak during the search; 561~MB during final predictor training.
\end{flushleft}
\end{table}

Uniform, Golden, and Gumbel have essentially identical computational costs.
Gumbel adds negligible cost because its selection is optimized jointly with
the predictor during training. Soft top-$K$ takes about 13\% longer and nearly
doubles the peak GPU memory, because its Soft and top-$K$ branches are both
evaluated before the selector is frozen. Agentic is the most expensive, at about 24~min, or 64\% above Uniform. The
extra time comes entirely from the search. Each candidate subset is
evaluated with a short 200-epoch training run of about 1~min, and a search
at $K=45$ uses about 9.6 such trials on average (8 to 12 across the five
searches). The final predictor training then costs the same as for the
other methods.


\subsection{Robustness to Measurement Noise}

\label{ssec:noise_robustness}
We tested sensitivity to measurement noise at $K = 45$ and $K = 2$. Following
Thombre et al.~\cite{thombre2026vasto}, Gaussian noise was added to the
selected view sinograms as
\begin{equation}
\widetilde{\mathbf{X}}_{\mathrm{noisy}}
=
\widetilde{\mathbf{X}}
+
\eta\,
\frac{\lVert \widetilde{\mathbf{X}} \rVert_2}
{\lVert \mathbf{Z} \rVert_2}\,
\mathbf{Z},
\qquad
Z_{ij} \sim \mathcal{N}(0,1),
\label{eq:relative_sinogram_noise}
\end{equation}
where $\mathbf{Z}$ is a noise matrix of the same size as the selected-view
sinogram $\widetilde{\mathbf{X}}$ (Eq.~\eqref{eq:mask}), with entries $Z_{ij}$
drawn independently from the standard normal distribution $\mathcal{N}(0,1)$,
and $\eta \in \{0.01, 0.05, 0.25\}$ sets the noise level relative to the
clean signal.

\begin{figure}[htbp]
  \centering
  \includegraphics[width=\textwidth]{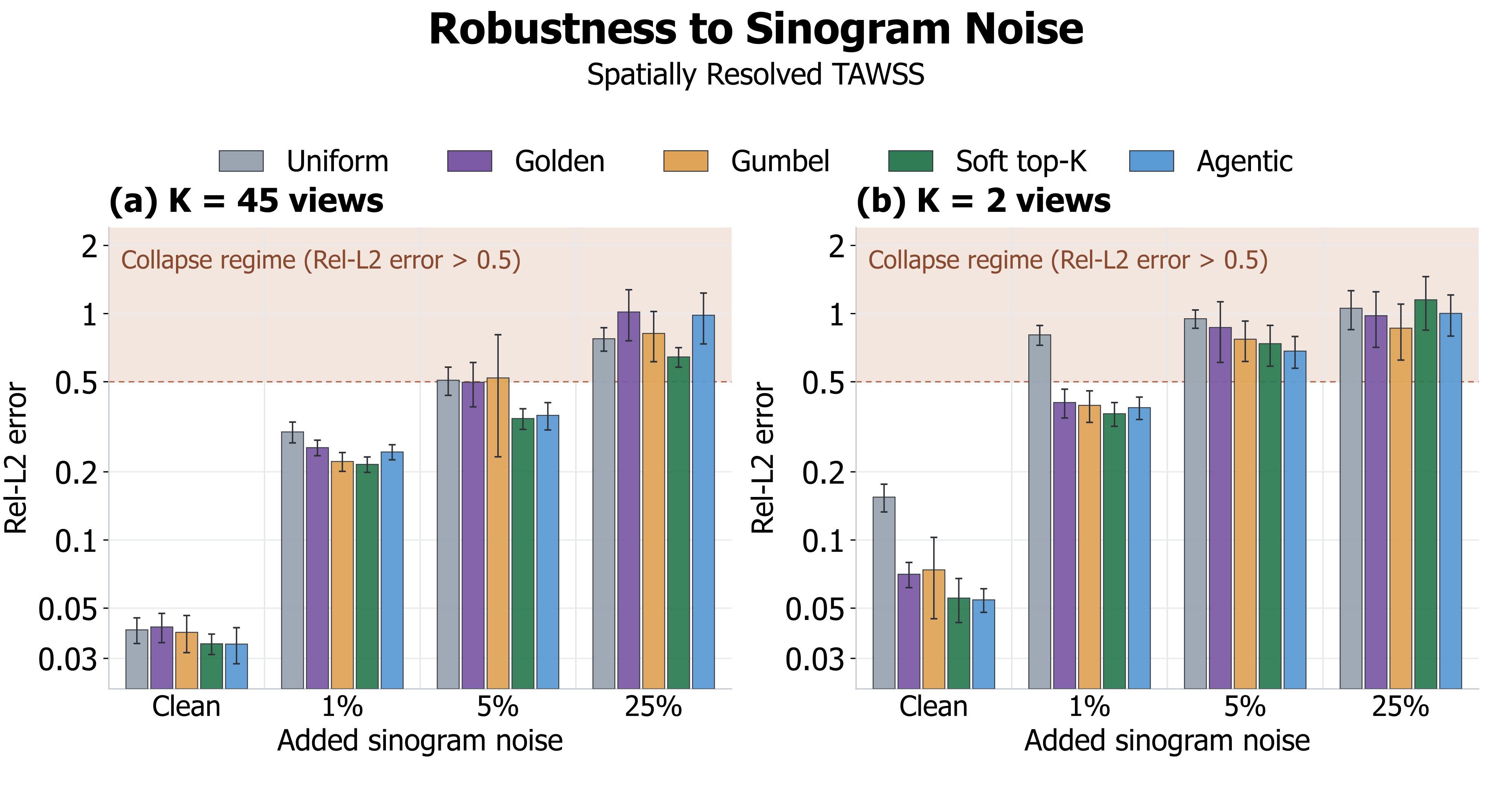}
  \caption{Robustness to additive sinogram noise for spatially resolved TAWSS.
  \textbf{(a)} $K = 45$ views. \textbf{(b)} $K = 2$ views. Bars are the mean
  and whiskers $\pm\,\sigma$ over five runs. The shaded region marks Rel-$L_2$ $> 0.5$, where the TAWSS predictor fails.}
  \label{fig:noise}
\end{figure}

All methods degrade quickly as noise is added (Fig.~\ref{fig:noise}). At low noise, the task-aware selections generally remain competitive, although the exact ranking differs from the clean case, and uniform sampling stays the worst, especially at $K = 2$. Selecting views for the task therefore still helps
under noise, but it does not make the prediction robust to it. At the highest noise level, and already at moderate noise for $K = 2$, all methods enter the shaded region of Fig.~\ref{fig:noise}, where the predictor collapses to a near-constant output and the ranking between methods is no longer meaningful. The sparsest budgets are the most sensitive, since with only one or two views there is little redundancy to average out the noise. Handling noise will likely require a dedicated approach such as a denoising step before the predictor or a noise model built into the training, and we discuss this further in Sec.~\ref{sec:discussion}.

\subsection{CTA Image Reconstruction}
\label{ssec:CT_reconstruct}

\begin{figure}[htbp]
  \centering
  \includegraphics[width=\textwidth]{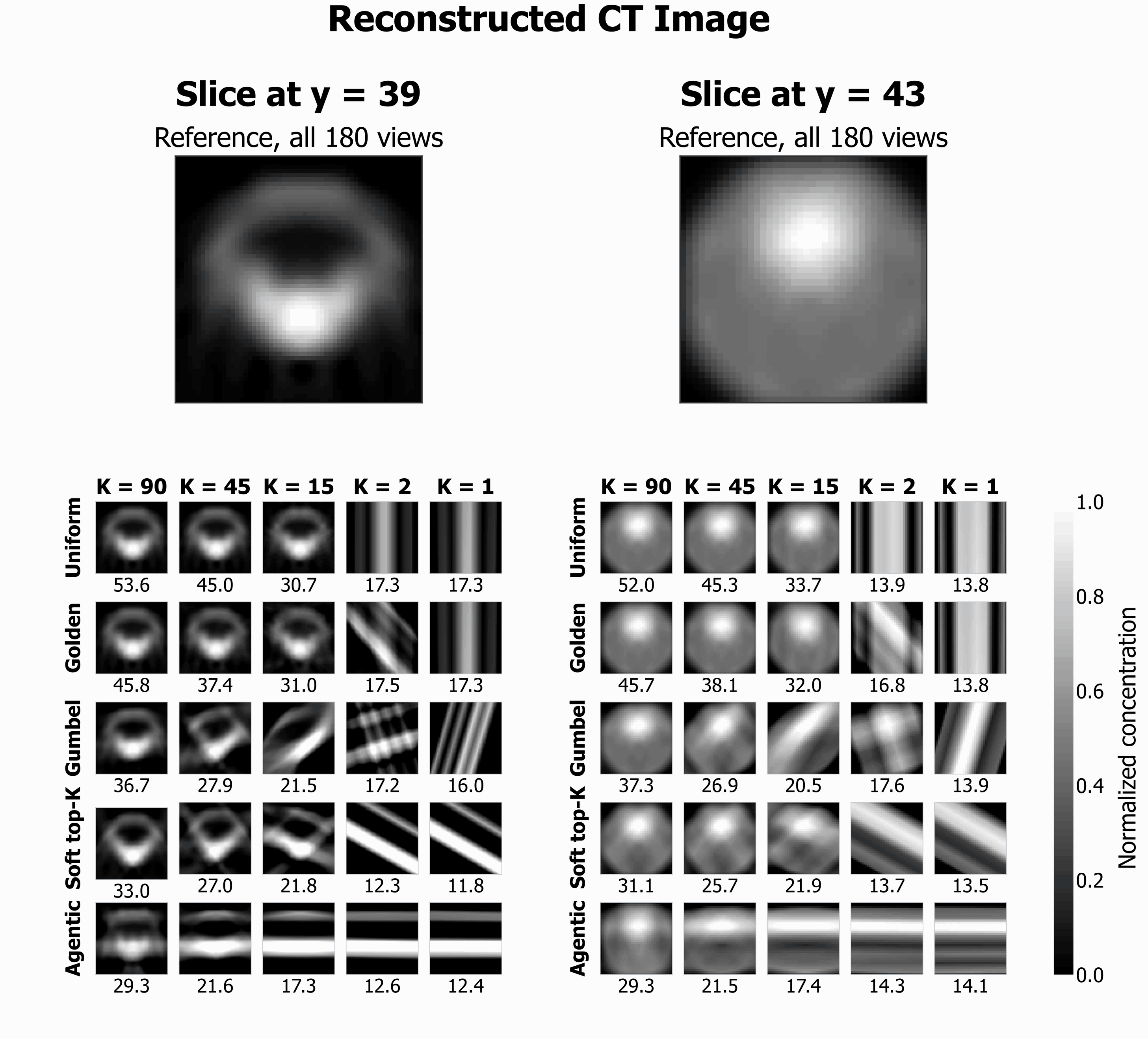}
  \caption{Filtered back-projection reconstructions from the mask each method selected for two $y$ slices of the same case at $t=2$~s.   Rows are the five view-selection methods and columns the view budget $K$. All panels are cropped to the object and the concentration is normalized to  $0$--$1$ within each slice. The number under each panel is the peak signal-to-noise ratio (PSNR) against that slice's reference.}
  \label{fig:sparse_recon}
\end{figure}

As a complementary analysis, we reconstructed the concentration field using FBP from the projection masks selected by each method. FBP is used here as a simple, transparent reference for how much image information each mask retains, and we should note that iterative reconstruction would improve absolute image quality (not performed here). Two slices from the same test case  were considered. For each method, the mask from one representative run was used, and the resulting sparse view reconstruction was compared with the corresponding full 180-view reference using the peak signal-to-noise ratio (PSNR). The reconstructed slices are shown in Fig.~\ref{fig:sparse_recon}.

At the denser budgets, the uniform mask produced the highest reconstruction quality, which contrasts with its performance for the task-aware TAWSS prediction problem. At $K=90$, the uniform mask achieved a PSNR of 53.6~dB for the first reconstructed slice, followed by the golden-angle mask at 45.8~dB. The learned masks produced lower PSNR values, ranging from 36.7~dB for Gumbel to 29.3~dB for Agentic. This difference indicates that the views selected to improve TAWSS prediction are not necessarily the same views that preserve conventional image reconstruction quality. At the sparser budgets, uniform sampling is no longer consistently the best. For example, Gumbel exceeds it at $K=2$ for the second slice (17.6 versus 13.9~dB).

The difference became more pronounced as the number of retained projections decreased. For the Agentic mask, the reconstructed image became strongly distorted by $K=45$. Below a PSNR of about 21~dB, the reconstructions no longer show a recognizable concentration field for any method. This comparison suggests that successful hemodynamic inference from a small number of projections does not require recovery of a visually accurate intermediate CTA image and suggests the need for joint learning to achieve both image reconstruction and accurate hemodynamic prediction.

\section{Discussion}
\label{sec:discussion}


In this study, we developed a framework that selects which sinogram views to acquire and predicts hemodynamics directly from the resulting sparse sinogram. Two fixed sampling baselines (uniform and golden-angle) were compared against two learned selection methods (Soft top-$K$ and Gumbel), and an agentic AI method that selects views through reasoning rather than gradient-based training. Our results showed that selecting the sinogram views for the task, either with the learned methods (Soft top-$K$ and Gumbel) or with the agentic method, improves TAWSS prediction over the fixed baselines, and that the improvement becomes more pronounced as the number of views decreases. For the spatially resolved target, the gain is largest at the sparsest budget ($K = 1$), while for the spatially averaged target the picture is different. Soft top-$K$ and Agentic give a similar gain only at the sparsest budgets, about 3.5 times lower error at $K = 1$, while at $K \ge 15$ the fixed baselines are as accurate or slightly better. This suggests that a scalar target such as the spatially averaged TAWSS does not require task-aware view selection unless the acquisition is planned to be very sparse. For most cases, the learned masks cluster around views perpendicular to the main arterial flow direction, concentrated where the contrast concentration changes most sharply, between the high-concentration jet at the sac and the surrounding low concentration region. These views were the most informative
for predicting TAWSS, but they degraded the quality of the reconstructed image, confirming that the best views for a downstream task are not necessarily the best views for image reconstruction. 

While obtaining multiple different selected subsets is aligned with previous work, the reasons for this non-uniqueness vary across applications. Ercan and Papadimitriou~\cite{ercan2021optimal} reported that optimal sensor placement problems can admit multiple local or global solutions, with different sensor configurations providing comparable information. Harman et al.~\cite{harman2024polytope} showed more formally that an optimal experimental design does not necessarily correspond to a unique configuration, and that multiple distinct designs can achieve the same optimality criterion. Similarly, Moreno-Salinas et al.~\cite{moreno2013optimal} demonstrated that different sensor geometries can satisfy the same optimality conditions when they provide equivalent relative measurement information. In medical imaging, it has been shown that different sampling patterns can provide useful complementary encoding rather than requiring one fixed set of measurements~\cite{breuer2006controlled}. More recently, Gautam et al.~\cite{gautam2026scan} showed that learned magnetic resonance imaging (MRI) sampling masks can depend on the initialization, reconstruction network, and loss function, leading to different optimized sampling patterns. Together, these studies support our observation that there may not be a uniquely identifiable best subset of views, and instead, multiple subsets can provide similar complementary information to the downstream neural network mapping and consequently achieve comparable prediction accuracy.

Our results show that, within the simulated setting studied here, one can predict TAWSS from only one or two carefully selected views, even when those views produce poor image reconstructions.  This suggests that hemodynamic prediction may require fewer measurements than recovering a detailed CTA image, offering a potential route to lower radiation exposure and direct prediction. In the case of MRI, it has been recently shown that highly undersampled $k$-space 4D flow MRI measurements with an appropriate sampling strategy could be used to obtain hemodynamic data and outperform traditional compressed sensing~\cite{locke2026parameter}. The benefit of view selection also depends on the target and budget, where fixed sampling was often sufficient for spatially averaged TAWSS at moderate budgets, while task-aware selection offered greater benefits for spatially resolved TAWSS as the budget decreased. From a practical perspective, the proposed view selection strategy could be translated to CT acquisition technology through angular tube-current modulation, hardware-aware selection of contiguous projection blocks, or dynamic beam attenuation/adaptive collimation. These approaches could reduce exposure at less informative angles while preserving measurements identified as important by the task-aware selector, providing feasible pathways toward implementing the learned acquisition patterns on real CT systems~\cite{mccollough2009strategies,kudo2013image,huck2019sheet,deak2009effects}. Furthermore, different masks achieved similar accuracy, suggesting that several effective subsets may exist and could provide freedom to choose a pattern that better satisfies hardware or acquisition constraints. 

The agentic approach adds flexibility by adapting its search based on previous results and combining strategies such as surrogate modeling, greedy selection, angular coverage, and refinement of earlier masks within a single search. Agentic AI had the lowest reported mean error at four budgets and tied Soft top-$K$ to four decimal places at \(K=90\) for spatially resolved TAWSS, supporting its potential for identifying effective subsets. This approach could extend to problems with practical constraints that are difficult to include in a loss function or differentiable optimization setting, such as placing sensors/views only at accessible locations, choosing experiments or CFD simulations under resource limits, and selecting model inputs that are costly to measure. In these settings, an agent could propose candidates, check feasibility, and improve its choices using measured performance without requiring gradients through the full evaluation process.


Several limitations should be acknowledged in the current study. First, the framework was evaluated using idealized synthetic data, which may not fully represent the variability and complexity encountered in patient-specific or clinical acquisitions. We also assumed that the contrast is quasi-static during each gantry rotation, whereas in practice the contrast evolves while the projections are acquired and the time-scale for evolution could be important. The proposed prediction framework also showed considerable sensitivity to measurement noise, particularly for highly sparse view masks. In addition, the present study considers only one CT acquisition geometry, namely fan-beam CT, and therefore the behavior of the selected masks under other acquisition designs remains unknown. Another limitation is that the view selection objective focuses entirely on the downstream hemodynamic prediction task while ignoring conventional image reconstruction quality. As a result, some masks that perform well for TAWSS prediction can produce poor reconstructed CTA images. The Agentic approach also introduces additional computational cost because multiple candidate masks must be proposed, evaluated, and refined during the search before the final mask is selected. Finally, the existence of multiple near-optimal subsets complicates interpretation of the selected views. A uniquely identifiable global optimum would make it easier to associate specific projection angles with their importance to the prediction task, whereas the presence of several similarly performing masks indicates that informative views may depend on combinations of complementary measurements.


Future work should focus on extending the framework toward more realistic and clinically relevant settings.  Since the current models are sensitive to measurement noise, future studies should also investigate denoising strategies, including learning a mapping from noisy sinograms to cleaner representations before downstream hemodynamic prediction and leveraging denoising methods used in deep learning~\cite{csala2024comparison}. To balance task-specific prediction with conventional imaging requirements, an auxiliary image-reconstruction loss could be incorporated into the optimization objective, allowing the selected mask to preserve both hemodynamic information and image quality. For the Agentic framework, future studies could compare different LLMs and reasoning strategies to determine how the choice of agent influences search efficiency, selected masks, and prediction performance. Hardware-aware agentic optimization could additionally enforce practical restrictions such as restricted angular regions, feasible tube-current modulation, or other scanner-specific constraints during the search.

\section{Conclusion}
\label{sec:conclusion}
We demonstrate that optimizing CTA projection views for downstream hemodynamic prediction can substantially improve performance under highly sparse acquisition. Agentic AI and Soft top-$K$ were especially effective for spatially resolved hemodynamic prediction, reducing the single-view error approximately 68--69\% compared with fixed sampling, while task-aware selection also provided clear benefits for spatially averaged TAWSS at very small view budgets. The results further show that multiple distinct view subsets can achieve similar accuracy and that views optimized for hemodynamic inference are not necessarily optimal for image reconstruction. Although validation on patient-specific and noisy clinical data is still needed, these findings support task-aware and agentic view selection as promising approaches for designing efficient CTA acquisitions (with fewer projections) for physical quantity prediction tasks.


\section*{Data Availability}
All code and data supporting this study will be made available at\\
\url{https://github.com/amir-cardiolab/Sparse_CT_AI}.

\section*{Acknowledgments}
\label{sec:acknowledgments}
This research was funded by the National Science Foundation (NSF) under Award No.~2205265.
\bibliographystyle{unsrt}
\bibliography{refs}

@article{zhou2017association,
  title={Association of wall shear stress with intracranial aneurysm rupture: systematic review and meta-analysis},
  author={Zhou, G. and Zhu, Y. and Yin, Y. and Su, M. and Li, M.},
  journal={Scientific Reports},
  volume={7},
  number={1},
  pages={5331},
  year={2017},
  publisher={Nature Publishing Group UK London}
}

@inproceedings{wang2023active,
  title={Active {CT} reconstruction with a learned sampling policy},
  author={Wang, C. and Shang, K. and Zhang, H. and Zhao, S. and Liang, D. and Zhou, S. K.},
  booktitle={Proceedings of the 31st ACM International Conference on Multimedia},
  pages={7226--7235},
  year={2023}
}

@article{yang2025learning,
  title={Learning Task-Specific Sampling Strategy for Sparse-View {CT} Reconstruction},
  author={Yang, L. and Huang, J. and Fang, Y. and Aviles-Rivero, A. I. and Sch{\"o}nlieb, C. and Zhang, D. and Yang, G.},
  journal={IEEE Transactions on Instrumentation and Measurement},
  year={2025},
  publisher={IEEE}
}

@article{meng2013high,
  title={High {WSS} or low {WSS}? Complex interactions of hemodynamics with intracranial aneurysm initiation, growth, and rupture: toward a unifying hypothesis},
  author={Meng, H. and Tutino, V. M. and Xiang, J. and Siddiqui, A.},
  journal={American journal of neuroradiology},
  year={2013},
  publisher={American Journal of Neuroradiology}
}

@article{thombre2026vasto,
  title={{VASTO}: Simultaneous recovery of vascular geometry and blood flow via differentiable topology optimization},
  author={Thombre, P. and Padhy, R. K. and D'Souza, R. M. and Suresh, K.},
  journal={arXiv preprint arXiv:2606.05487},
  year={2026}
}

@inproceedings{palenstijn2013astra,
  title={The {ASTRA} tomography toolbox},
  author={Palenstijn, W. J. and Batenburg, K. J. and Sijbers, J.},
  booktitle={13th International Conference on Computational and Mathematical Methods in Science and Engineering, CMMSE},
  volume={4},
  pages={1139--1145},
  year={2013}
}

@article{elhadidy2026sle,
  title={{SLE-FNO}: Single-layer extensions for task-agnostic continual learning in Fourier neural operators},
  author={Elhadidy, M. and D’Souza, R. M. and Arzani, A.},
  journal={Physica Scripta},
  volume={101},
  number={21},
  pages={216002},
  year={2026},
  publisher={IOP Publishing}
}

@article{elhadidy2026wall,
  title={Wall Shear Stress Reconstruction from Concentration: Differentiable Physics and Physics-Informed Neural Networks},
  author={Elhadidy, M. and Viknesh, S. and D'Souza, R. M. and Arzani, A.},
   journal = {Physics of Fluids},
    volume = {38},
    number = {8},
    pages = {081908},
  year={2026}
}

@article{kato1999application,
  title={Application of three-dimensional {CT} angiography {(3D-CTA)} to cerebral aneurysms},
  author={Kato, Y. and Sano, H. and Katada, K. and Ogura, Y. and Hayakawa, M. and Kanaoka, N. and Kanno, T.},
  journal={Surgical Neurology},
  volume={52},
  number={2},
  pages={113--122},
  year={1999},
  publisher={Elsevier}
}

@article{jin2026predicting,
  title={Predicting Intracranial Aneurysm Rupture Risk and Intervention Outcomes Using Denoising-Enhanced {CTA}},
  author={Jin, H. and Chen, L. and Zhu, T. and Chu, G. and Ma, L. and Liang, G. and Zou, Z. and Yu, C.},
  journal={American Journal of Neuroradiology},
  volume={47},
  number={5},
  pages={1250--1258},
  year={2026},
  publisher={American Society of Neuroradiology}
}

@article{khatak2026radiation,
  title={Radiation Risks of Computed Tomography: Estimating Cancer Incidence from Diagnostic Imaging},
  author={Khatak, A. and Kumar, S. and Alam, M. M. and Saha, S. and Verma, A. and Sharma, S. and Singh, D. K.},
  journal={INTERNATIONAL JOURNAL OF ADVANCES IN SIGNAL AND IMAGE SCIENCES},
  pages={1--14},
  year={2026}
}

@article{smith2025projected,
  title={Projected lifetime cancer risks from current computed tomography imaging},
  author={Smith-Bindman, R. and Chu, P. W. and Azman Firdaus, H. and Stewart, C. and Malekhedayat, M. and Alber, S. and Bolch, W. E. and Mahendra, M. and Berrington de Gonz{\'a}lez, A. and Miglioretti, D. L.},
  journal={JAMA Internal Medicine},
  volume={185},
  number={6},
  pages={710--719},
  year={2025}
}

@article{zhang2018sparse,
  title={A sparse-view {CT} reconstruction method based on combination of DenseNet and deconvolution},
  author={Zhang, Z. and Liang, X. and Dong, X. and Xie, Y. and Cao, G.},
  journal={IEEE transactions on medical imaging},
  volume={37},
  number={6},
  pages={1407--1417},
  year={2018},
  publisher={IEEE}
}

@article{lee2018deep,
  title={Deep-neural-network-based sinogram synthesis for sparse-view {CT} image reconstruction},
  author={Lee, H. and Lee, J. and Kim, H. and Cho, B. and Cho, S.},
  journal={IEEE Transactions on Radiation and Plasma Medical Sciences},
  volume={3},
  number={2},
  pages={109--119},
  year={2018},
  publisher={IEEE}
}

@inproceedings{kohler2004projection,
  title={A projection access scheme for iterative reconstruction based on the golden section},
  author={Kohler, T.},
  booktitle={IEEE Symposium Conference Record Nuclear Science 2004.},
  volume={6},
  pages={3961--3965},
  year={2004},
  organization={IEEE}
}

@article{wang2026deep,
  title={Deep reinforcement learning for optimising angle selection and dose allocation in {CT} reconstruction},
  author={Wang, T. and Pelt, D. M. and Lucka, F. and van Leeuwen, T. and Batenburg, K. J.},
  journal={Applied Mathematics in Science and Engineering},
  volume={34},
  number={1},
  pages={2705107},
  year={2026},
  publisher={Taylor \& Francis}
}

@article{wang2025dynamic,
  title={Dynamic angle selection in {X-Ray} {CT}: A reinforcement learning approach to optimal stopping},
  author={Wang, T. and Lucka, F. and Pelt, D. M. and Batenburg, K. J. and van Leeuwen, T.},
  journal={arXiv preprint arXiv:2503.12688},
  year={2025}
}

@article{shen2020learning,
  title={Learning to scan: A deep reinforcement learning approach for personalized scanning in {CT} imaging},
  author={Shen, Z. and Wang, Y. and Wu, D. and Yang, X. and Dong, B.},
  journal={arXiv preprint arXiv:2006.02420},
  year={2020}
}

@inproceedings{yang2022learning,
  title={Learning projection views for sparse-view {CT} reconstruction},
  author={Yang, L. and Ge, R. and Feng, S. and Zhang, D.},
  booktitle={Proceedings of the 30th ACM International Conference on Multimedia},
  pages={2645--2653},
  year={2022}
}

@article{mccollough2009strategies,
  title={Strategies for reducing radiation dose in {CT}},
  author={McCollough, C. H. and Primak, A. N. and Braun, N. and Kofler, J. and Yu, L. and Christner, J.},
  journal={Radiologic Clinics of North America},
  volume={47},
  number={1},
  pages={27},
  year={2009}
}

@article{kudo2013image,
  title={Image reconstruction for sparse-view {CT} and interior {CT}—introduction to compressed sensing and differentiated backprojection},
  author={Kudo, H. and Suzuki, T. and Rashed, E. A.},
  journal={Quantitative Imaging in Medicine and Surgery},
  volume={3},
  number={3},
  pages={147},
  year={2013}
}

@article{huck2019sheet,
  title={Sheet-based dynamic beam attenuator--A novel concept for dynamic fluence field modulation in x-ray {CT}},
  author={Huck, S. M. and Fung, G. S. and Parodi, K. and Stierstorfer, K.},
  journal={Medical Physics},
  volume={46},
  number={12},
  pages={5528--5537},
  year={2019},
  publisher={Wiley Online Library}
}

@article{deak2009effects,
  title={Effects of adaptive section collimation on patient radiation dose in multisection spiral {CT}},
  author={Deak, P. D. and Langner, O. and Lell, M. and Kalender, W. A.},
  journal={Radiology},
  volume={252},
  number={1},
  pages={140--147},
  year={2009},
  publisher={Radiological Society of North America}
}

@inproceedings{wu2021fractal,
  title={Fractal autoencoders for feature selection},
  author={Wu, X. and Cheng, Q.},
  booktitle={Proceedings of the AAAI Conference on Artificial Intelligence},
  volume={35},
  number={12},
  pages={10370--10378},
  year={2021}
}

@article{bengio2013estimating,
  title={Estimating or propagating gradients through stochastic neurons for conditional computation},
  author={Bengio, Y. and L{\'e}onard, N. and Courville, A.},
  journal={arXiv preprint arXiv:1308.3432},
  year={2013}
}

@inproceedings{balin2019concrete,
  title={Concrete autoencoders: Differentiable feature selection and reconstruction},
  author={Bal{\i}n, M. F. and Abid, A. and Zou, J.},
  booktitle={International Conference on Machine Learning},
  pages={444--453},
  year={2019},
  organization={PMLR}
}

@inproceedings{maddison2017a,
  journal = {Proceedings of the international conference on learning Representations},
  booktitle = {},
  pages = {1-20},
  publisher = {International Conference on Learning Representations},
  school = {},
  title = {The concrete distribution: A continuous relaxation of discrete random variables},
  volume = {},
  author = {Maddison, C. and Mnih, A. and Teh, Y.},
  editor = {},
  year = {2017},
  organizer = {},
  series = {}
}

@inproceedings{jang2017categorical,
title={Categorical Reparameterization with Gumbel-Softmax},
author={E. Jang and S. Gu and B. Poole},
booktitle={International Conference on Learning Representations},
year={2017},
url={https://openreview.net/forum?id=rkE3y85ee}
}

@inproceedings{yang2024large,
  title={Large language models as optimizers},
  author={Yang, C. and Wang, X. and Lu, Y. and Liu, H. and Le, Q. V. and Zhou, D. and Chen, X.},
  booktitle={International Conference on Learning Representations},
  volume={2024},
  pages={12028--12068},
  year={2024}
}

@article{li2020fourier,
  title={Fourier neural operator for parametric partial differential equations},
  author={Li, Z. and Kovachki, N. and Azizzadenesheli, K. and Liu, B. and Bhattacharya, K. and Stuart, A. and Anandkumar, A.},
  journal={arXiv preprint arXiv:2010.08895},
  year={2020}
}

@article{kovachki2023neural,
  title={Neural operator: Learning maps between function spaces with applications to pdes},
  author={Kovachki, N. and Li, Z. and Liu, B. and Azizzadenesheli, K. and Bhattacharya, K. and Stuart, A. and Anandkumar, A.},
  journal={Journal of Machine Learning Research},
  volume={24},
  number={89},
  pages={1--97},
  year={2023}
}

@article{li2023fourier,
  title={Fourier neural operator with learned deformations for {PDEs} on general geometries},
  author={Li, Z. and Huang, D. Z. and Liu, B. and Anandkumar, A.},
  journal={Journal of Machine Learning Research},
  volume={24},
  number={388},
  pages={1--26},
  year={2023}
}

@inproceedings{flouris2025localized,
  title={Localized {FNO} for Spatiotemporal Hemodynamic Upsampling in Aneurysm {MRI}},
  author={Flouris, K. and Halter, M. and Lee, Y. Y. and Castonguay, S. and Jacobs, L. and Dirix, P. and Nestmann, J. and Kozerke, S. and Konukoglu, E.},
  booktitle={International Workshop on Reconstruction and Imaging Motion Estimation},
  pages={65--75},
  year={2025},
  organization={Springer}
}

@inproceedings{hu2018squeeze,
  title={Squeeze-and-excitation networks},
  author={Hu, J. and Shen, L. and Sun, G.},
  booktitle={Proceedings of the IEEE Conference on Computer Vision and Pattern Recognition},
  pages={7132--7141},
  year={2018}
}

@article{joseph1980view,
  title={View sampling requirements in fan beam computed tomography},
  author={Joseph, P. M. and Schulz, R. A.},
  journal={Medical Physics},
  volume={7},
  number={6},
  pages={692--702},
  year={1980},
  publisher={Wiley Online Library}
}

@book{kak2001principles,
  title={Principles of computerized tomographic imaging},
  author={Kak, A. C. and Slaney, M.},
  year={2001},
  publisher={SIAM}
}

@article{zeng2022filtered,
  title={Filtered Back-Projection Reconstruction with Non-Uniformly Under-Sampled Projections},
  author={Zeng, G. L. and Zeng, M.},
  journal={Archives in Biomedical Engineering \& Biotechnology},
  volume={7},
  number={1},
  pages={000652},
  year={2022}
}

@article{frikel2013characterization,
  title={Characterization and reduction of artifacts in limited angle tomography},
  author={Frikel, J. and Quinto, E. T.},
  journal={Inverse Problems},
  volume={29},
  number={12},
  pages={125007},
  year={2013},
  publisher={IOP Publishing}
}

@article{zhang2024review,
  title={A review of deep learning methods for denoising of medical low-dose {CT} images},
  author={Zhang, J. and Gong, W. and Ye, L. and Wang, F. and Shangguan, Z. and Cheng, Y.},
  journal={Computers in Biology and Medicine},
  volume={171},
  pages={108112},
  year={2024},
  publisher={Elsevier}
}

@article{chen2024deep,
  title={Deep learning-based algorithms for low-dose {CT} imaging: A review},
  author={Chen, H. and Li, Qiuxia and Zhou, Lazhen and Li, Fangzuo},
  journal={European Journal of Radiology},
  volume={172},
  pages={111355},
  year={2024},
  publisher={Elsevier}
}

@article{hsieh2013recent,
  title={Recent advances in {CT} image reconstruction},
      author={Hsieh, J. and Nett, B. and Yu, Z. and Sauer, K. and Thibault, J. B. and Bouman, C. A.},
  journal={Current Radiology Reports},
  volume={1},
  number={1},
  pages={39--51},
  year={2013},
  publisher={Springer}
}

@article{koetzier2023deep,
  title={Deep learning image reconstruction for {CT}: technical principles and clinical prospects},
  author={Koetzier, L. R. and Mastrodicasa, D. and Szczykutowicz, T. P. and van der Werf, N. R. and Wang, A. S. and Sandfort, V. and van der Molen, A. J. and Fleischmann, D. and Willemink, M. J.},
  journal={Radiology},
  volume={306},
  number={3},
  pages={e221257},
  year={2023},
  publisher={Radiological Society of North America}
}

@article{kulathilake2023review,
  title={A review on deep learning approaches for low-dose computed tomography restoration},
  author={Kulathilake, K. S. Hemantha and Abdullah, N. A. and Sabri, A. Q. M. and Lai, K. W.},
  journal={Complex \& Intelligent Systems},
  volume={9},
  number={3},
  pages={2713--2745},
  year={2023},
  publisher={Springer}
}

@article{enjilela2019cubic,
  title={Cubic-spline interpolation for sparse-view {CT} image reconstruction with filtered backprojection in dynamic myocardial perfusion imaging},
  author={Enjilela, E. and Lee, T.-Y. and Wisenberg, G. and Teefy, P. and Bagur, R. and Islam, A. and Hsieh, J. and So, A.},
  journal={Tomography},
  volume={5},
  number={3},
  pages={300},
  year={2019}
}

@article{chen2008prior,
  title={Prior image constrained compressed sensing ({PICCS}): a method to accurately reconstruct dynamic {CT} images from highly undersampled projection data sets},
  author={Chen, G.-H. and Tang, J. and Leng, S.},
  journal={Medical Physics},
  volume={35},
  number={2},
  pages={660--663},
  year={2008},
  publisher={Wiley Online Library}
}

@article{ernst2023sinogram,
  title={Sinogram upsampling using Primal--Dual UNet for undersampled {CT} and radial {MRI} reconstruction},
  author={Ernst, P. and Chatterjee, S. and Rose, G. and Speck, O. and N{\"u}rnberger, A.},
  journal={Neural Networks},
  volume={166},
  pages={704--721},
  year={2023},
  publisher={Elsevier}
}

@article{wu2019computationally,
  title={Computationally efficient deep neural network for computed tomography image reconstruction},
  author={Wu, D. and Kim, K. and Li, Q.},
  journal={Medical Physics},
  volume={46},
  number={11},
  pages={4763--4776},
  year={2019},
  publisher={Wiley Online Library}
}

@inproceedings{wang2022dudotrans,
  title={{DuDoTrans}: Dual-domain transformer for sparse-view {CT} reconstruction},
  author={Wang, C. and Shang, K. and Zhang, H. and Li, Q. and Zhou, S. K.},
  booktitle={International Workshop on Machine Learning for Medical Image Reconstruction},
  pages={84--94},
  year={2022},
  organization={Springer}
}

@article{li2022ddptransformer,
  title={{DDPTransformer}: dual-domain with parallel transformer network for sparse view {CT} image reconstruction},
  author={Li, R. and Li, Q. and Wang, H. and Li, S. and Zhao, J. and Yan, Q. and Wang, L.},
  journal={IEEE Transactions on Computational Imaging},
  volume={8},
  pages={1101--1116},
  year={2022},
  publisher={IEEE}
}

@inproceedings{zhou2025spatiotemporal,
  title={Spatiotemporal-aware neural fields for dynamic {CT} reconstruction},
  author={Zhou, Q. and Ye, Y. and Cai, Z.},
  booktitle={Proceedings of the AAAI Conference on Artificial Intelligence},
  volume={39},
  number={10},
  pages={10834--10842},
  year={2025}
}

@article{zhou2025non,
  title={Non-local guided neural fields for {4D CT} reconstruction},
  author={Zhou, Q. and Ye, Y. and Liu, Z. and Liu, C. and Cai, Z.},
  journal={IEEE Transactions on Circuits and Systems for Video Technology},
  year={2025},
  publisher={IEEE}
}

@article{liu2024evolution,
  title={Evolution of heuristics: Towards efficient automatic algorithm design using large language model},
  author={Liu, F. and Tong, X. and Yuan, M. and Lin, X. and Luo, F. and Wang, Z. and Lu, Z. and Zhang, Q.},
  journal={arXiv preprint arXiv:2401.02051},
  year={2024}
}

@article{romera2024mathematical,
  title={Mathematical discoveries from program search with large language models},
  author={Romera-Paredes, B. and Barekatain, M. and Novikov, A. and Balog, M. and Kumar, M. P. and Dupont, E. and Ruiz, F. J. and Ellenberg, J. S. and Wang, P. and Fawzi, O. and others},
  journal={Nature},
  volume={625},
  number={7995},
  pages={468--475},
  year={2024},
  publisher={Nature Publishing Group UK London}
}

@article{yuksekgonul2024textgrad,
  title={Textgrad: Automatic" differentiation" via text},
  author={Yuksekgonul, M. and Bianchi, F. and Boen, J. and Liu, S. and Huang, Z. and Guestrin, C. and Zou, J.},
  journal={arXiv preprint arXiv:2406.07496},
  year={2024}
}

@inproceedings{yuan2026unleashing,
  title={Unleashing {LLMs} in bayesian optimization: Preference-guided framework for scientific discovery},
  author={Yuan, X. and Chen, Z. and Zhang, J. and Xiong, H. and Ye, N. and Li, Y. and Gu, Q.},
  booktitle={The Fourteenth International Conference on Learning Representations},
  year={2026}
}

@article{boiko2023autonomous,
  title={Autonomous chemical research with large language models},
  author={Boiko, D. A. and MacKnight, R. and Kline, B. and Gomes, G.},
  journal={Nature},
  volume={624},
  number={7992},
  pages={570--578},
  year={2023},
  publisher={Nature Publishing Group UK London}
}

@article{toscano2026graft,
  title={{GRAFT-ATHENA}: Self-improving agentic teams for autonomous discovery and evolutionary numerical algorithms},
  author={Toscano, J. D. and Chai, Z. and Karniadakis, G. E.},
  journal={arXiv preprint arXiv:2605.11117},
  year={2026}
}

@article{jiang2026agenticsciml,
  title={Agenticsciml: Collaborative multi-agent systems for emergent discovery in scientific machine learning},
  author={Jiang, Q. and Karniadakis, G.},
  journal={npj Artificial Intelligence},
  year={2026},
  publisher={Nature Publishing Group UK London}
}

@article{maier2026agentic,
  title={Agentic Autoresearch for CT Reconstruction},
  author={Maier, A. and Kachelriess, L. and Bayer, S. and Huang, Y. and Xia, Y. and Simpson, A. and Zaiss, M.},
  journal={arXiv preprint arXiv:2607.22824},
  year={2026}
}

@article{zaiss2026agentic,
  title={Agentic {MR} sequence development: leveraging {LLMs} with MR skills for automatic physics-informed sequence development},
  author={Zaiss, M. and Aly, A. and Endres, J. and Dornstetter, T. and Weinm{\"u}ller, S. and Maier, A.},
  journal={arXiv preprint arXiv:2604.13282},
  year={2026}
}

@article{lardo2015estimating,
  title={Estimating coronary blood flow using {CT} transluminal attenuation flow encoding: Formulation, preclinical validation, and clinical feasibility},
  author={Lardo, A. C. and Rahsepar, A. A. and Seo, J. H. and Eslami, P. and Korley, F. and Kishi, S. and Abd, T. and Mittal, R. and George, R. T.},
  journal={Journal of Cardiovascular Computed Tomography},
  volume={9},
  number={6},
  pages={559--566},
  year={2015},
  publisher={Elsevier}
}

@article{eslami2015computational,
  title={Computational study of computed tomography contrast gradients in models of stenosed coronary arteries},
  author={Eslami, P. and Seo, J. H. and Rahsepar, A. A. and George, R. and Lardo, A. C. and Mittal, R.},
  journal={Journal of Biomechanical Engineering},
  volume={137},
  number={9},
  pages={091002},
  year={2015},
  publisher={American Society of Mechanical Engineers}
}

@article{raissi2020hidden,
  title={Hidden fluid mechanics: Learning velocity and pressure fields from flow visualizations},
  author={Raissi, M. and Yazdani, A. and Karniadakis, G. E.},
  journal={Science},
  volume={367},
  number={6481},
  pages={1026--1030},
  year={2020},
  publisher={American Association for the Advancement of Science}
}

@inproceedings{datta2026resolution,
  title={Resolution-Agnostic Neural Operators for Multi-rate Sparse-View {CT}},
  author={Datta, A. and Wang, J. and Aali, A. and Anandkumar, A.},
  booktitle={European Conference on Computer Vision},
  pages={243--262},
  year={2026},
  organization={Springer}
}

@article{brunton2013optimal,
  title={Optimal sensor placement and enhanced sparsity for classification},
  author={Brunton, B. W. and Brunton, S. L. and Proctor, J. L. and Kutz, J. N.},
  journal={arXiv preprint arXiv:1310.4217},
  year={2013}
}

@article{semaan2017optimal,
  title={Optimal sensor placement using machine learning},
  author={Semaan, R.},
  journal={Computers \& Fluids},
  volume={159},
  pages={167--176},
  year={2017},
  publisher={Elsevier}
}

@article{nav2026hierarchical,
  title={A hierarchical machine learning framework for real-time reconstruction of urban wind fields from sparse sensor networks},
  author={Nav, F. M. and Snaiki, R. and Ding, F.},
  journal={Building and Environment},
  pages={114875},
  year={2026},
  publisher={Elsevier}
}

@article{mahmoudi2021story,
  title={The story of wall shear stress in coronary artery atherosclerosis: biochemical transport and mechanotransduction},
  author={Mahmoudi, M. and Farghadan, A. and McConnell, D. R. and Barker, A. J. and Wentzel, J. J. and Budoff, M. J. and Arzani, A.},
  journal={Journal of Biomechanical Engineering},
  volume={143},
  number={4},
  pages={041002},
  year={2021},
  publisher={American Society of Mechanical Engineers}
}

@article{ercan2021optimal,
  title={Optimal sensor placement for reliable virtual sensing using modal expansion and information theory},
  author={Ercan, T. and Papadimitriou, C.},
  journal={Sensors},
  volume={21},
  number={10},
  pages={3400},
  year={2021},
  publisher={MDPI}
}

@article{harman2024polytope,
  title={The polytope of optimal approximate designs: extending the selection of informative experiments},
  author={Harman, R. and Filov{\'a}, L. and Rosa, S.},
  journal={Statistics and Computing},
  volume={34},
  number={6},
  pages={211},
  year={2024},
  publisher={Springer}
}

@article{moreno2013optimal,
  title={Optimal sensor placement for multiple target positioning with range-only measurements in two-dimensional scenarios},
  author={Moreno-Salinas, D. and Pascoal, A. M and Aranda, J.},
  journal={Sensors},
  volume={13},
  number={8},
  pages={10674--10710},
  year={2013},
  publisher={Molecular Diversity Preservation International (MDPI)}
}

@article{breuer2006controlled,
  title={Controlled aliasing in volumetric parallel imaging ({2D CAIPIRINHA})},
  author={Breuer, F. A. and Blaimer, M. and Mueller, M. F. and Seiberlich, N. and Heidemann, R. M. and Griswold, M. A. and Jakob, P. M.},
  journal={Magnetic Resonance in Medicine: An Official Journal of the International Society for Magnetic Resonance in Medicine},
  volume={55},
  number={3},
  pages={549--556},
  year={2006},
  publisher={Wiley Online Library}
}

@article{gautam2026scan,
  title={Scan-adaptive {MRI} undersampling using neighbor-based optimization (suno)},
  author={Gautam, S. and Li, A. and Seiberlich, N. and Fessler, J. A. and Ravishankar, S.},
  journal={IEEE Transactions on Computational Imaging},
  year={2026},
  publisher={IEEE}
}

@article{sharma2019analytic,
  title={Analytic reconstruction of a two-dimensional velocity field from an observed diffusive scalar},
  author={Sharma, A. and Rypina, I. I. and Musgrave, R. and Haller, G.},
  journal={Journal of Fluid Mechanics},
  volume={871},
  pages={755--774},
  year={2019},
  publisher={Cambridge University Press}
}

@inproceedings{sechidis2019stability,
  title={On the stability of feature selection in the presence of feature correlations},
  author={Sechidis, K. and Papangelou, K. and Nogueira, S. and Weatherall, J. and Brown, G.},
  booktitle={Joint European Conference on Machine Learning and Knowledge Discovery in Databases},
  pages={327--342},
  year={2019},
  organization={Springer}
}

@inproceedings{hamer2021robust,
  title={Robust Selection Stability Estimation in Correlated Spaces},
  author={Hamer, V. and Dupont, P.},
  booktitle={Joint European Conference on Machine Learning and Knowledge Discovery in Databases},
  pages={446--461},
  year={2021},
  organization={Springer}
}

@article{csala2024comparison,
  title={A comparison of machine learning methods for recovering noisy and missing {4D flow MRI} data},
  author={Csala, H. and Amili, O. and D'Souza, R. M. and Arzani, A.},
  journal={International Journal for Numerical Methods in Biomedical Engineering},
  volume={40},
  number={11},
  pages={e3858},
  year={2024},
  publisher={Wiley Online Library}
}

@article{sliwicka2023dynamic,
  title={Dynamic myocardial {CT} perfusion imaging—state of the art},
  author={Sliwicka, O. and Sechopoulos, I. and Baggiano, A. and Pontone, G. and Nijveldt, R. and Habets, J.},
  journal={European Radiology},
  volume={33},
  number={8},
  pages={5509--5525},
  year={2023},
  publisher={Springer}
}

@article{vu2020evaluation,
  title={Evaluation of multislice inputs to convolutional neural networks for medical image segmentation},
  author={Vu, M. H. and Grimbergen, G. and Nyholm, T. and L{\"o}fstedt, T.},
  journal={Medical Physics},
  volume={47},
  number={12},
  pages={6216--6231},
  year={2020},
  publisher={Wiley Online Library}
}

@article{isensee2021nnu,
  title={{nnU-Net}: a self-configuring method for deep learning-based biomedical image segmentation},
  author={Isensee, F. and Jaeger, P. F. and Kohl, S. A. and Petersen, J. and Maier-Hein, K. H.},
  journal={Nature methods},
  volume={18},
  number={2},
  pages={203--211},
  year={2021},
  publisher={Nature Publishing Group US New York}
}

@inproceedings{christ2016automatic,
  title={Automatic liver and lesion segmentation in {CT} using cascaded fully convolutional neural networks and {3D} conditional random fields},
  author={Christ, P. Ferdinand and Elshaer, M. E. A. and Ettlinger, F. and Tatavarty, S. and Bickel, M. and Bilic, P. and Rempfler, M. and Armbruster, M. and Hofmann, F. and D’Anastasi, M. and others},
  booktitle={International Conference on Medical Image Computing and Computer-Assisted Intervention},
  pages={415--423},
  year={2016},
  organization={Springer}
}

@article{roth2018application,
  title={An application of cascaded {3D} fully convolutional networks for medical image segmentation},
  author={Roth, H. R. and Oda, H. and Zhou, X. and Shimizu, N. and Yang, Y. and Hayashi, Y. and Oda, M. and Fujiwara, M. and Misawa, K. and Mori, K.},
  journal={Computerized Medical Imaging and Graphics},
  volume={66},
  pages={90--99},
  year={2018},
  publisher={Elsevier}
}

@inproceedings{zhou2017fixed,
  title={A fixed-point model for pancreas segmentation in abdominal {CT} scans},
  author={Zhou, Y. and Xie, L. and Shen, W. and Wang, Y. and Fishman, E. K. and Yuille, A. L.},
  booktitle={International Conference on Medical Image Computing and Computer-Assisted Intervention},
  pages={693--701},
  year={2017},
  organization={Springer}
}

@article{han2017automatic,
  title={Automatic liver lesion segmentation using a deep convolutional neural network method},
  author={Han, X.},
  journal={arXiv preprint arXiv:1704.07239},
  year={2017}
}

@article{li2018h,
  title={{H-DenseUNet}: hybrid densely connected {UNet} for liver and tumor segmentation from {CT} volumes},
  author={Li, X. and Chen, H. and Qi, X. and Dou, Q. and Fu, C.-W. and Heng, P.-A.},
  journal={IEEE Transactions on Medical Imaging},
  volume={37},
  number={12},
  pages={2663--2674},
  year={2018},
  publisher={IEEE}
}

@article{locke2026parameter,
  title={Parameter estimation in blood flow models from highly undersampled k-space magnetic resonance imaging data},
  author={L{\"o}cke, M. and van Ooij, P. and Bertoglio, C.},
  journal={Biomechanics and Modeling in Mechanobiology},
  volume={25},
  number={4},
  pages={74},
  year={2026},
  publisher={Springer}
}

@article{guo2026sinogram,
  title={Sinogram-based flow estimation in computed tomography using a physics-informed neural network: Impact of gantry rotation speed, {X-ray} fluence and pulsed acquisition on accuracy},
  author={Guo, J. and Khurana, G. S. and Grande, A. G. and Alamo, J. C. del and Contijoch, F.},
  journal={Medical Physics},
  volume={53},
  number={6},
  pages={e70519},
  year={2026},
  publisher={Wiley Online Library}
}

@article{withers2021x,
  title={{X-ray} computed tomography},
  author={Withers, P. J. and Bouman, C. and Carmignato, S. and Cnudde, V. and others},
  journal={Nature Reviews Methods Primers},
  volume={1},
  number={1},
  pages={18},
  year={2021},
  publisher={Nature Publishing Group UK London}
}

@article{lee2026agentic,
  title={Agentic {AI} for Computational Fluid Dynamics: A Review},
  author={Lee, K. and Kim, I.},
  journal={Applied Mechanics},
  volume={7},
  number={4},
  pages={81},
  year={2026},
  publisher={Multidisciplinary Digital Publishing Institute}
}

\end{document}